\documentclass[10pt,twocolumn,letterpaper]{article}

\usepackage[pagenumbers]{cvpr} 

\usepackage{wrapfig}

\usepackage{duckuments}
\usepackage{multirow}

\definecolor{cvprblue}{rgb}{0.21,0.49,0.74}
\usepackage[pagebackref,breaklinks,colorlinks,allcolors=cvprblue]{hyperref}

\newcommand{\projname}{\textit{DiffusionBrowser}}

\newcommand{\xzeropred}{\(\mathbf{x}_0\)\nobreakdash-pred}

\newcommand\blfootnote[1]{%
  \begingroup
  \renewcommand\thefootnote{}\footnote{#1}%
  \addtocounter{footnote}{-1}%
  \endgroup
}

\def\paperID{****} 
\def\confName{CVPR}
\def\confYear{2026}

\title{DiffusionBrowser: Interactive Diffusion Previews via Multi-Branch Decoders}

\author{
    Susung Hong$^{1,*}$
    \quad\quad Chongjian Ge$^2$
    \quad\quad Zhifei Zhang$^2$
    \quad\quad Jui-Hsien Wang$^2$ \\
   $^1$ University of Washington \quad $^2$ Adobe Research \\
    \url{https://susunghong.github.io/DiffusionBrowser}
}

\begin{document}

\twocolumn[
\begin{@twocolumnfalse}
\maketitle
\begin{center}
    \centering
    \captionsetup{type=figure}
    \includegraphics[trim={0 0 0 4mm},clip,width=.98\linewidth]{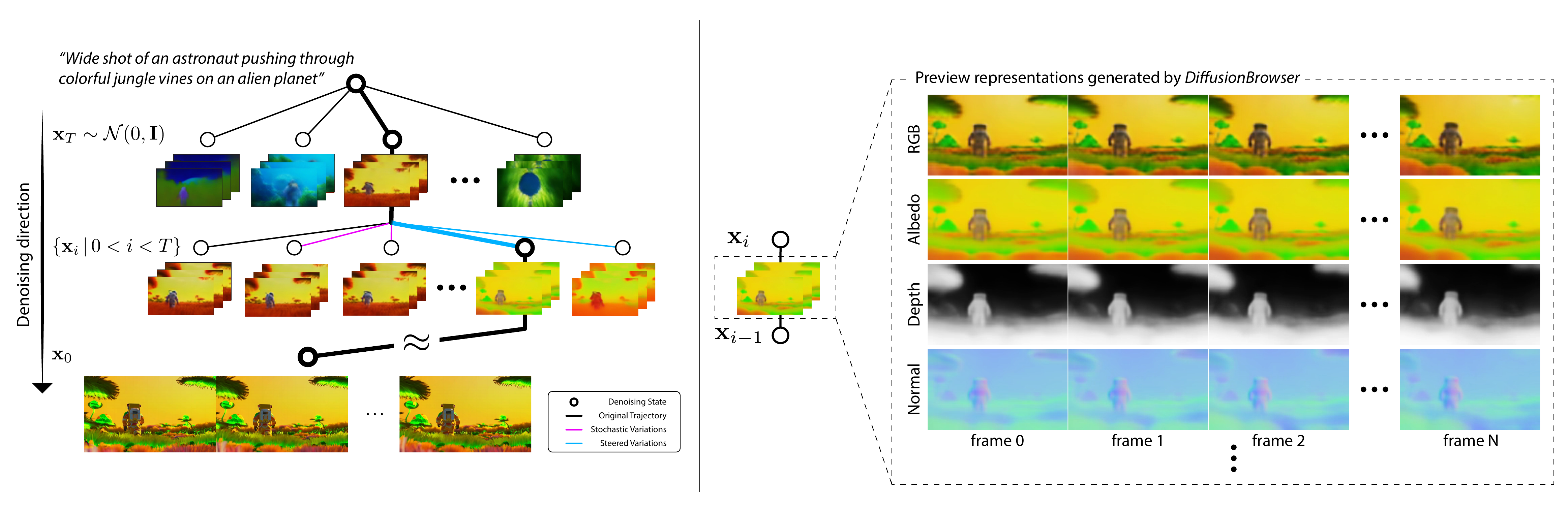}
    \vspace{-4mm}
    \captionof{figure}{
    \textbf{\projname{}} is a plug-and-play model that enables interactive previews anywhere in the multi-step diffusion process, which allows users to make decisions about whether to continue denoising or modify prompts. In addition, \projname{} provides multiple variation generation mechanisms that guide users to explore the creative generation space in a tree-like structure. Our novel, efficient multi-branch decoder architecture preserves the full capacity of the base model, and can generate rich multi-modal previews for each timestep in $<1$s, adding negligible overhead at inference time.
    }
    \label{fig:teaser}
\end{center}
\end{@twocolumnfalse}
]
\blfootnote{* Work done during an internship at Adobe.}

\begin{abstract}
Video diffusion models have revolutionized generative video synthesis, but they are imprecise, slow, and can be opaque during generation---keeping users in the dark for a prolonged period. In this work, we propose \projname{}, a model-agnostic, lightweight decoder framework that allows users to interactively generate previews at any point (timestep or transformer block) during the denoising process. Our model can generate multi-modal preview representations that include RGB and scene intrinsics at more than 4$\times$ real-time speed (less than 1 second for a 4-second video) that convey consistent appearance and motion to the final video. With the trained decoder, we show that it is possible to interactively guide the generation at intermediate noise steps via stochasticity reinjection and modal steering, unlocking a new control capability. Moreover, we systematically probe the model using the learned decoders, revealing how scene, object, and other details are composed and assembled during the otherwise black-box denoising process.
\end{abstract}

\section{Introduction}
Modern video diffusion models possess remarkable capabilities to generate vivid depictions of diverse scenes. However, two fundamental challenges remain for practical deployment: 1) \textit{limited controllability}, which results from the inherent stochasticity of the diffusion processes, and 2) \textit{slow generation speed}, which restricts iterative creation and efficient workflows. Recent work has explored adding various control mechanisms such as camera or object controls, and conditional input modalities such as edge or depth to make video generation more predictable. Another body of work focuses on improving training and inference efficiency, for example, via distillation, mixture-of-experts, autoregressive models, sparse attention, etc. However, while these methods mitigate the two issues, they come with consequences. For example, distillation can cause mode collapse and quality degradation, and adding extra input conditioning like depth can change base model quality and complicate training and inference setups. Even if these techniques work perfectly, diffusion models are still naturally stochastic, and hence some amount of uncertainty remains.

To address these limitations, we propose \projname{}, a model-agnostic, lightweight framework to provide users with consistent \emph{previews} at any given point in the denoising process (block-wise or denoising step-wise), and do so without compromising the model's full capacity and with negligible overhead. The previews allow users to terminate irrelevant generations early and save inference resources. 

Drawing inspiration from traditional graphics rendering pipelines, we designed \projname{} to be able to preview auxiliary intrinsic channels such as albedo, depth, and surface normals on top of RGB pixels. We show that these intrinsics emerge early in the generation process and can be decoded using a carefully designed multi-branch, modality-optimized decoder. Our preview heads provide a more complete 3D preview of the final generation and can be used to understand the inner workings of diffusion models to provide insights on various blocks and timesteps.

An additional benefit of the preview decoders is to unlock a new form of generation control by steering the denoising trajectories at sample time, allowing users to interact with the vast diversity provided by the generation model. Because \projname{} surfaces semantically meaningful signals such as coarse layout, motion direction, and appearance at early timesteps, users can intervene before the model commits to a non-ideal path. We demonstrate this by showing examples of color, depth, and normal steering at various branching points, providing a decision tree-like, interactive generation capability (see Figure~\ref{fig:teaser}).

Our main contributions are summarized below:
\begin{itemize}
\item We introduce \projname{}, a lightweight framework that provides \emph{previews} during video diffusion, enabling early termination without degrading fidelity.
\item Our method preserves full-generation quality, supports rapid iteration, and is fully plug-and-play.
\item Inspired by classical rendering pipelines and by the emergence of intrinsics early in denoising, we produce rich previews of RGB and intrinsic channels (albedo, depth, normals) through a multi-branch, multi-loss predictor.
\item The preview heads enable novel, interactive generation control by steering denoising trajectories using early semantic signals (layout, motion, appearance), allowing users to guide outcomes at branching points.
\end{itemize}
\section{Related Work}
\label{sec:relwork}

\paragraph{Efficiency-Based Methods.}
Many studies have improved the efficiency of video diffusion models by reducing sampling steps or simplifying model architectures. Distillation-based approaches~\citep{yin2025slow,lin2025diffusion,yang2025towards} compress multi-step denoising into fewer steps but often suffer from quality degradation and reduced output diversity. Cascaded diffusion models, such as FlashVideo~\citep{zhang2025flashvideo}, adopt a coarse-to-fine generation strategy; however, each stage still requires full-step inference, limiting practical speedup. Autoregressive frameworks~\citep{lin2025autoregressive,huang2025self,deng2024autoregressive} generate frames sequentially to mitigate long-horizon error accumulation but are not suitable for real-time preview during generation. Other directions explore efficiency through techniques such as mixture-of-experts~\citep{ganjdanesh2024mixture} and sparse attention~\citep{zhang2025vsa,chen2025sana,zhan2025bidirectional}, aiming to reduce computational cost. Despite these advances, existing methods often involve complex training pipelines or modify the model's capacity. In contrast, our approach remains model-agnostic and preserves full generation fidelity while enabling fast previews and user steering without altering the underlying diffusion model.
\vspace{-20pt}

\paragraph{Diffusion Features.}
Diffusion models generate high-quality images and videos through iterative denoising~\citep{ho2020denoising, song2020score, rombach2022high}. Past research has focused on their internal representations to understand how semantics, structure, and style are encoded, from older U-Net-based architectures~\citep{kwon2022diffusion} to more recent transformer-based ones~\citep{ahn2025fine}. Cross-attention maps reveal how text tokens align with visual features, enabling semantic control and editing~\citep{hertz2022prompt, chefer2023attend,helbling2025conceptattention}, though they do not fully explain objects spontaneously included in the scene. Meanwhile, studies of self-attention and intermediate states show these components carry rich structural information independent of text conditioning~\citep{hong2023improving, hong2024smoothed, ahn2024self, hyung2025spatiotemporal, geyer2023tokenflow}. Applications of diffusion features include image-to-image translation~\cite{zhang2021plug}, correspondence~\cite{tang2023emergent,stracke2025cleandift,zhang2023tale,luo2023diffusion,luo2024readout}, and zero-shot video generation~\cite{khachatryan2023text2video,huang2023free,hong2023direct2v}. In this paper, we define the novel task of preview generation and steering that can serve as a new tool to analyze video diffusion features.
\vspace{-20pt}

\paragraph{Generative Models and Scene Intrinsics.}
Prior work has shown that image generative models, such as GANs and diffusion models, encode geometry and shading cues, enabling applications such as depth estimation and relighting~\citep{kim2024depth, chen2023beyond, du2023generative, zhang2025scaling, zeng2024rgb, el2024probing}. These findings suggest that, despite being trained solely on 2D images, generative models implicitly learn both inverse and forward rendering processes. Our paper builds on this work to predict intrinsics from intermediate diffusion features.
\vspace{-20pt}

\paragraph{Post-Training Alignment and Reinforcement Learning.}
Previous work has explored training-time finetuning~\cite{fan2023dpokreinforcementlearningfinetuning,black2024trainingdiffusionmodelsreinforcement} or inference-time adaptation of diffusion models~\cite{he2023manifoldpreservingguideddiffusion,jain2025diffusiontreesamplingscalable} using reinforcement learning. In particular, \citet{jain2025diffusiontreesamplingscalable} recently proposed casting generation as a tree search problem and showed how a reward function can be used to guide sampling. However, the reward function used in that work, such as the aesthetic score, is accessible only after roll-out to obtain a clean sample. In contrast, our work focuses on efficiently generating multi-modal previews at intermediate nodes, with which users can then steer the generation. In other words, we circumvent the expensive reward evaluation and directly align with user preference by design, and therefore can complement this existing work.
\begin{figure*}[t]
\centering
\includegraphics[width=1.0\linewidth]{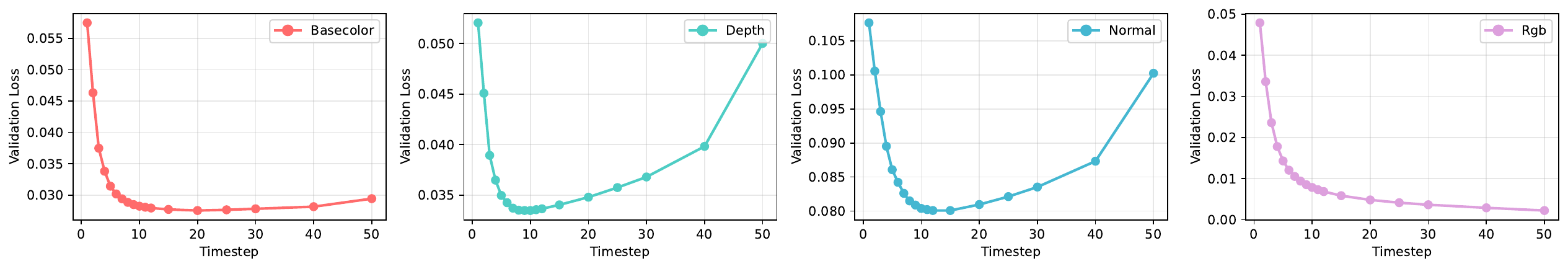}
\includegraphics[width=1.0\linewidth]{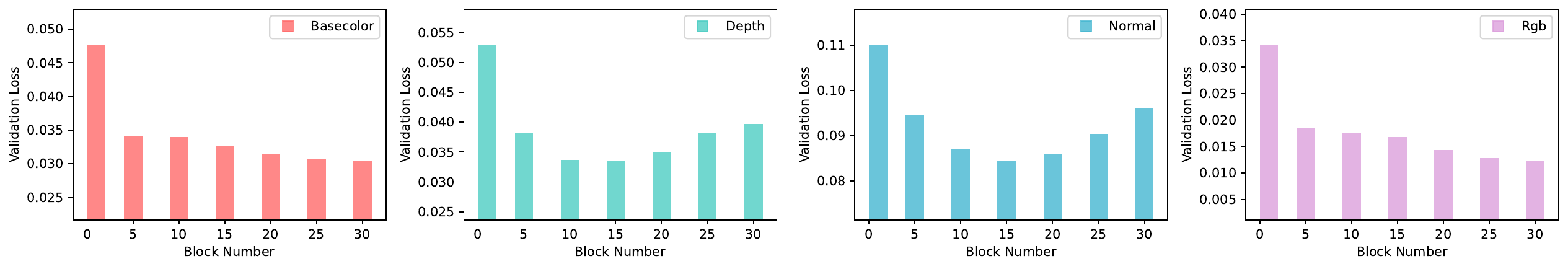}
\vspace{-20pt}
\caption{\textbf{Linear probing} results for scene intrinsics (base color, depth, and normals) and RGB with respect to timesteps and blocks. We use a single linear layer with MSE loss (cosine loss for normals). Predictive power from features to scene intrinsics saturates quickly both across blocks and across timesteps---around the 5th–15th of 50 timesteps and the 10th–20th of 30 blocks. Depth and normals are more easily predicted at earlier stages, whereas RGB prediction quality increases monotonically with both layer depth and timestep. Similar patterns are observed in nonlinear analyses (see the supplementary material), confirming that scene intrinsics can be captured quickly.}
\label{fig:linear-probing}
\vspace{-10pt}
\end{figure*}

\section{Background}
\subsection{Diffusion Models}
Diffusion models~\cite{ho2020denoising, song2020score} learn to synthesize data by reversing a gradual noising process. The noising process for an image $\mathbf{x}$ over time $t \in [0, 1]$ is:
\begin{equation}
d\mathbf{x} = \mathbf{f}(\mathbf{x}, t)dt + g(t)d\mathbf{w},
\end{equation}
where $\mathbf{f}$ and $g$ are drift and diffusion functions, and $d\mathbf{w}$ is a standard Wiener process. Instead of sampling from the reverse-time stochastic differential equations, one can equivalently solve their associated deterministic ordinary differential equations, which yield the same marginal distributions under suitable conditions. This perspective enables flow-matching approaches~\cite{lipman2022flow, liu2022flow}. Flow-matching samplers frame generative modeling as learning a vector field $\mathbf{v}_\theta(\mathbf{x}_t, t)$ whose trajectories satisfy:
\begin{equation}
\frac{d\mathbf{x}_t}{dt} = \mathbf{v}_\theta(\mathbf{x}_t, t),
\end{equation}
and whose induced flow matches the data distribution at $t=0$. In practice, we simulate this continuous dynamics via a discrete sequence of $T$ steps, updating $\mathbf{x}_t$ iteratively according to the learned vector field.

\subsection{Scene Intrinsics}
\label{sec:background_scene_intrinsics}

Scene intrinsics is a multi-channel representation used to describe the set of geometric, shading, and lighting information from images, which has been long studied in computer vision and graphics communities~\cite{saito1990comprehensible,bell14intrinsic}. Since the decomposition is mathematically under-constrained, recent studies have turned to diffusion models to leverage their stochasticity to sample the possible solution space~\cite{zeng2024rgb,liang2025diffusion,kocsis2025intrinsix,kocsis2024intrinsicimage}; given an image or video, these diffusion models are trained to predict scene intrinsics, including geometric channels such as depth and surface normals, and appearance channels such as base color (albedo), roughness, and metallic. 

\section{Interactive Preview Generation}
\label{sec:spontaneous}

We discuss the appropriate preview representation and introduce our framework, \projname{}, that enables interactive preview generation.

\subsection{Previewing With Scene Intrinsics}
\label{sec:preview_scene_intrinsics}

We seek efficient and semantically meaningful preview representations that satisfy the following two conditions:
\begin{enumerate}
    \item[(1)] Human users with the preview can determine what will be generated in the full-fidelity generation.
    \item[(2)] Diffusion models can generate the representation at earlier stages of denoising in order to make previews efficient.
\end{enumerate}
Note that the two goals are contradictory: the best, most perceptually consistent preview for humans is the final output, which is the last stage of a reasonable diffusion model (otherwise, the model is unnecessarily large). We therefore approach the problem by finding a representation that compromises (1) as long as key perceptual factors such as appearance and motion can be depicted. Scene intrinsics offer an appealing choice in that if we discard irradiance (lighting), then the rest of the channels such as albedo and depth are lower frequency signals consisting of larger, colorful patches, yet object boundaries and scene compositions are still visible (see Figure~\ref{fig:teaser} for an example). In addition, previous work has suggested early emergence of low frequency structures in RGB~\cite{wu2024freeinit}, which offers a good prospect for satisfying (2).

\vspace{-10pt}
\paragraph{Indeed, intrinsic scene representations emerge early in the denoising process.} We demonstrate this by training a set of linear probes for the scene intrinsics. Specifically, given a transformer-based diffusion model with $N_b$ blocks and a denoising schedule that involves $N_t$ steps, we attach linear projection layers, each at a distinct block $b$ and timestep $t$, to predict target intrinsic maps $\mathbf{y}_{t,b} \in \{\mathbf{b}, \mathbf{d}, \mathbf{n}, \mathbf{r}, \mathbf{m}, \mathbf{c}\}$, for base color, depth, surface normals, material roughness, material metallicity, and color RGB, respectively. The results in Figure~\ref{fig:linear-probing} clearly show that intrinsics emerge early blockwise, but especially stepwise, supporting our thesis that these semantic features can be useful in early-step preview generation. More training details can be found in~\S\ref{sec:implementation_details}.

\subsection{Multi-Branch Preview Predictor}
\label{sec:mb_predictor}

We leverage the results from the linear probing experiment to determine the semantically meaningful diffusion features and aim to build a better predictor in this section.

\vspace{-10pt}
\paragraph{Naive Predictor.} One simple approach to improve the linear probing results is to use a deeper model. Since we want a lightweight decoder and are limited by data scale, we choose 3D convolutional layers with channel-specific losses. Specifically, given a prediction $\hat{\mathbf{y}} = [\hat{\mathbf{b}}, \hat{\mathbf{d}}, \hat{\mathbf{n}}, \hat{\mathbf{m}}, \hat{\mathbf{r}}, \hat{\mathbf{c}}]$, the per-channel losses are written as
\begin{align}
\mathcal{L}_{\mathbf{o}} &= \| \hat{\mathbf{o}} - \mathbf{o} \|_1, \quad \mathbf{o} \in \{\mathbf{b}, \mathbf{c}\}, \\
\mathcal{L}_{\mathbf{s}} &= \| \hat{\mathbf{s}} - \mathbf{s} \|_1, \quad \mathbf{s} \in \{\mathbf{d}, \mathbf{m}, \mathbf{r}\}, \\
\mathcal{L}_{\mathbf{n}} &= 1 - \frac{\hat{\mathbf{n}} \cdot \mathbf{n}}{\| \hat{\mathbf{n}} \|_2 \| \mathbf{n} \|_2}.
\end{align}
The complete loss function for the naive predictor is the sum of all channels, $\mathcal{L}_n = \sum_{j \in \mathcal{I}} \mathcal{L}_j,$ where $\mathcal{I} = \{\mathbf{b}, \mathbf{d}, \mathbf{n}, \mathbf{m}, \mathbf{r}, \mathbf{c}\}$.

\vspace{-10pt}
\paragraph{Superposition Problem.} We noticed that with the naive predictor above, the results can contain certain blurry parts, especially at high motion or spatially complex patches (e.g., around fast-moving hands). We hypothesize that this is similar to the hallucination problem (e.g., generated hand images containing unseen six fingers~\cite{aithal2024hallucinations}), but one that happens at intermediate parts of the denoising trajectory caused by superimposed spatiotemporal uncertainty. Specifically, the estimated posterior mean $\hat{\mathbf{x}}_0 = \mathbb{E}[\mathbf{x}_0 | \mathbf{x}_t]$ generated by models trained with mean squared error learns a smoothed approximation of the true score function and can temporarily push samples toward low-density regions between modes of the data distribution. This is especially problematic at noisy timesteps because, intuitively, the conditional likelihood $p(\mathbf{x}_t | \mathbf{x}_0) = \mathcal{N}((1-\sigma_t)\mathbf{x}_0, \sigma_t^2 \mathbf{I})$ broadens significantly for large $t$, and thus the conditional posterior $p(\mathbf{x}_0 | \mathbf{x}_t)$ can be highly multimodal. In other words, the blurred parts are likely superpositions of plausible nearby trajectories supported by multimodal data (e.g., one trajectory with a hand moving up, the other down, and the sample is superimposed to manifest as a blurry patch at high $t$). If this does not get sufficiently resolved near $t=0$, it becomes hallucinated samples (e.g., six-finger hands).

\begin{figure}[!ht]
\centering
\includegraphics[width=1.0\linewidth]{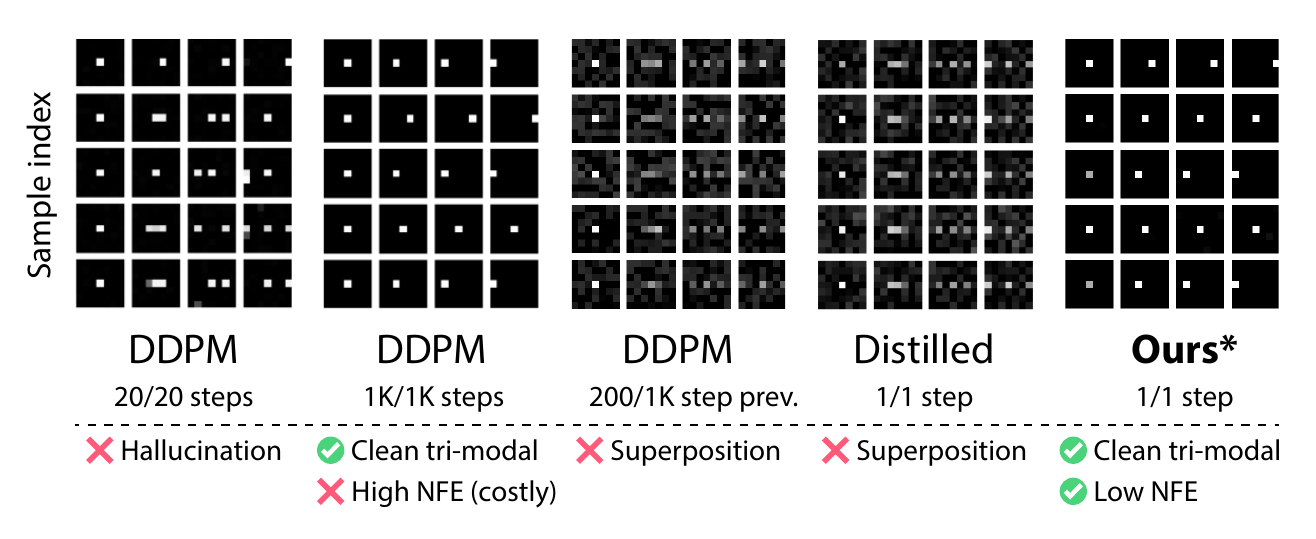}
\vspace{-20pt}
\caption{\textbf{Diffusion models trained on a toy 4-frame tri-modal dataset} reveal severe hallucination and superposition problems at low NFE settings (low total number of steps, distilled few-step model, or early preview) for models trained with MSE. In contrast, our multi-branch decoder architecture correctly produces a clean tri-modal distribution and remains artifact-free. For our results, the samples are randomly extracted from different branches, which learned to favor different modes in the data.
}
\vspace{-10pt}
\label{fig:toy}
\end{figure}

\vspace{-10pt}
\paragraph{Toy problem illustrates the superposition problem.} 
\label{sec:toy_problem}
To test our analysis, we constructed a tri-modal dataset containing 4-frame videos of a single white dot moving left, right, or remaining stationary (see Figure~\ref{fig:toy}). We then trained a diffusion model on this dataset using a 4-layer DiT (with roughly 0.3M parameters) with standard $\epsilon$-parameterization~\cite{ho2020denoising}. Once we trained the model, we additionally distilled it with consistency distillation~\cite{song2023consistency}. Running inference on these toy models leads to the following observations supporting our hypothesis: 1) the superposition problem occurs in $\hat{\mathbf{x}}_0$ at earlier timesteps such as 200 of 1,000 steps (see the third panel in Figure~\ref{fig:toy}); 2) using coarser timestep discretization such as 20-step DDPM and consistency-distilled models with 1 step exhibits more severe superposition problems, which create multiple high-intensity dots at the same time or completely remove the dots, which never occurs in the toy training videos (see the first and fourth panels in Figure~\ref{fig:toy}). This motivates us to address states in superposition when predicting clean signals.

\begin{figure}[!ht]
\centering
\includegraphics[width=0.94\linewidth]{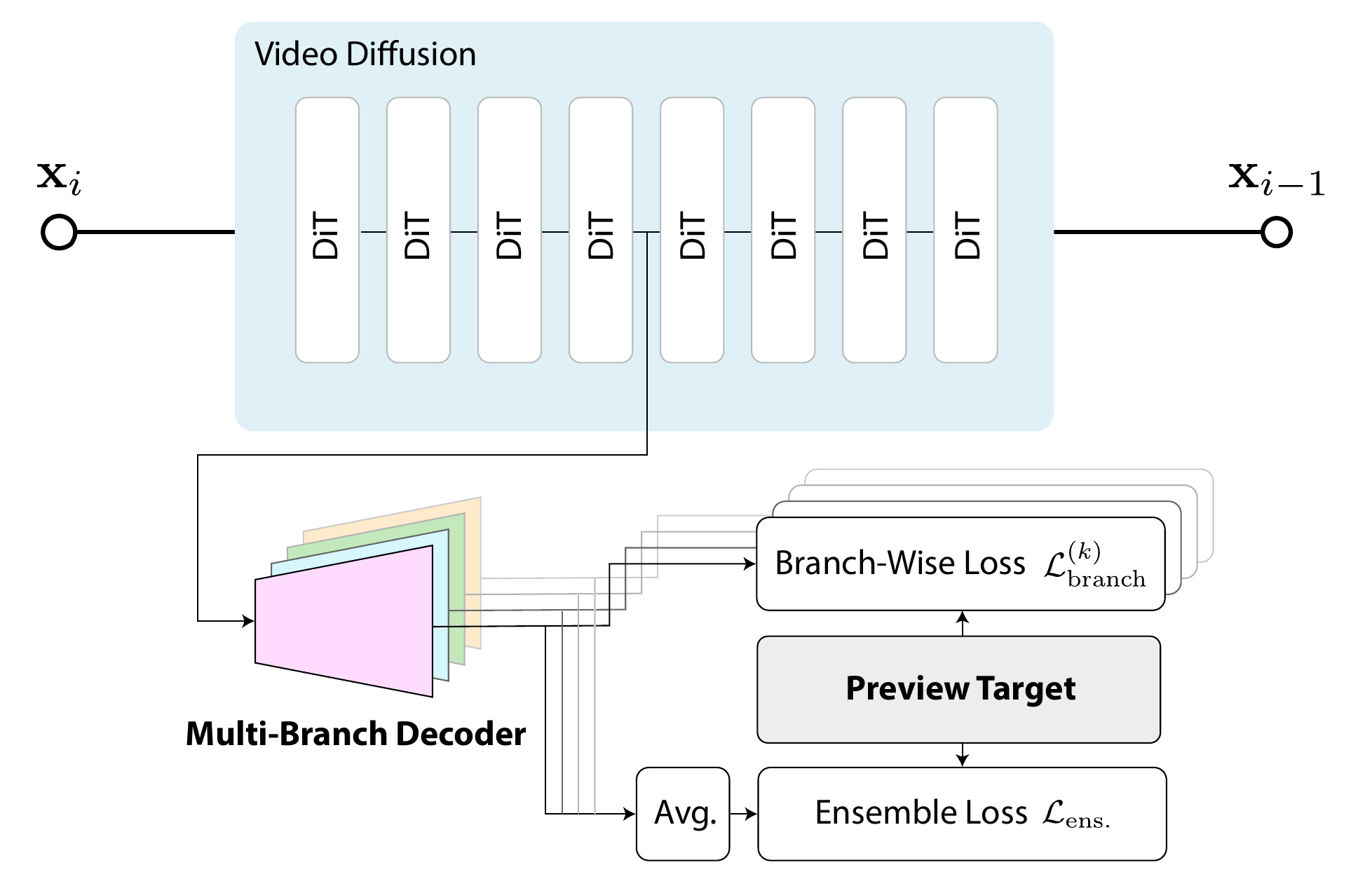}
\vspace{-10pt}
\caption{\textbf{Our multi-branch multi-loss decoder} is trained with intermediate diffusion features. Grounded by branch-wise loss and an aggregated ensemble loss, it is designed to reduce the superposition problem.}
\vspace{-7pt}
\label{fig:pipeline}
\end{figure}

\vspace{-10pt}
\paragraph{Multi-Branch Multi-Loss Predictor}
\label{sec:predictor_design}
We mitigate the superposition problem with a multi-branch decoding architecture (MB); see Figure~\ref{fig:pipeline} for an illustration. Instead of a single deterministic head, we introduce $K$ independent decoders $\{\mathcal{D}_k\}_{k=1}^K$, each predicting intrinsic maps $\hat{\mathbf{y}}_k = \mathcal{D}_k(\mathbf{f}_{t,b})$. Their ensemble average
\begin{align}
\hat{\mathbf{y}}_{\text{ens.}} = \frac{1}{K} \sum_{k=1}^{K} \mathcal{D}_k(\mathbf{f}_{t,b})
\end{align}
is trained jointly with individual heads. The total loss combines individual branch losses $\mathcal{L}_{n}^{(k)}$ (reusing the loss from the naive predictor) with an ensemble loss:
\begin{align}
\mathcal{L}_{\text{ens.}} &= \mathbb{E}\!\left[\left\| \hat{\mathbf{y}}_{\text{ens.}}^{\backslash \mathbf{n}} - \mathbf{y}^{\backslash \mathbf{n}} \right\|_2^2\right] + 1 - \frac{\hat{\mathbf{n}}_{\text{ens.}} \cdot \mathbf{n}}{\|\hat{\mathbf{n}}_{\text{ens.}}\|_2 \|\mathbf{n}\|_2}, \\
\mathcal{L}_{\text{total}} &= \lambda_{\text{ens.}}\mathcal{L}_{\text{ens.}} + \sum_{k=1}^{K} \mathcal{L}_{n}^{(k)},
\end{align}
where $\mathbf{y}^{\backslash \mathbf{n}} = [\mathbf{b}, \mathbf{d}, \mathbf{m}, \mathbf{r}, \mathbf{c}]$ denotes all intrinsic components except normals $\mathbf{n}$, which require a directional loss. With a proper branch-wise loss, each branch prediction represents a possible mode in the data distribution. We find that using a mode-seeking loss (e.g., LPIPS) for $\mathcal{L}_{n}$ together with multi-branching resolves the superposition problem even with a single NFE (see Figure~\ref{fig:toy} and \S\ref{sec:supp_toy_problem}) and also ensures that the mean of the branches closely matches the ground-truth mean of the data.

\begin{figure}[!ht]
\centering
\includegraphics[width=1.0\linewidth]{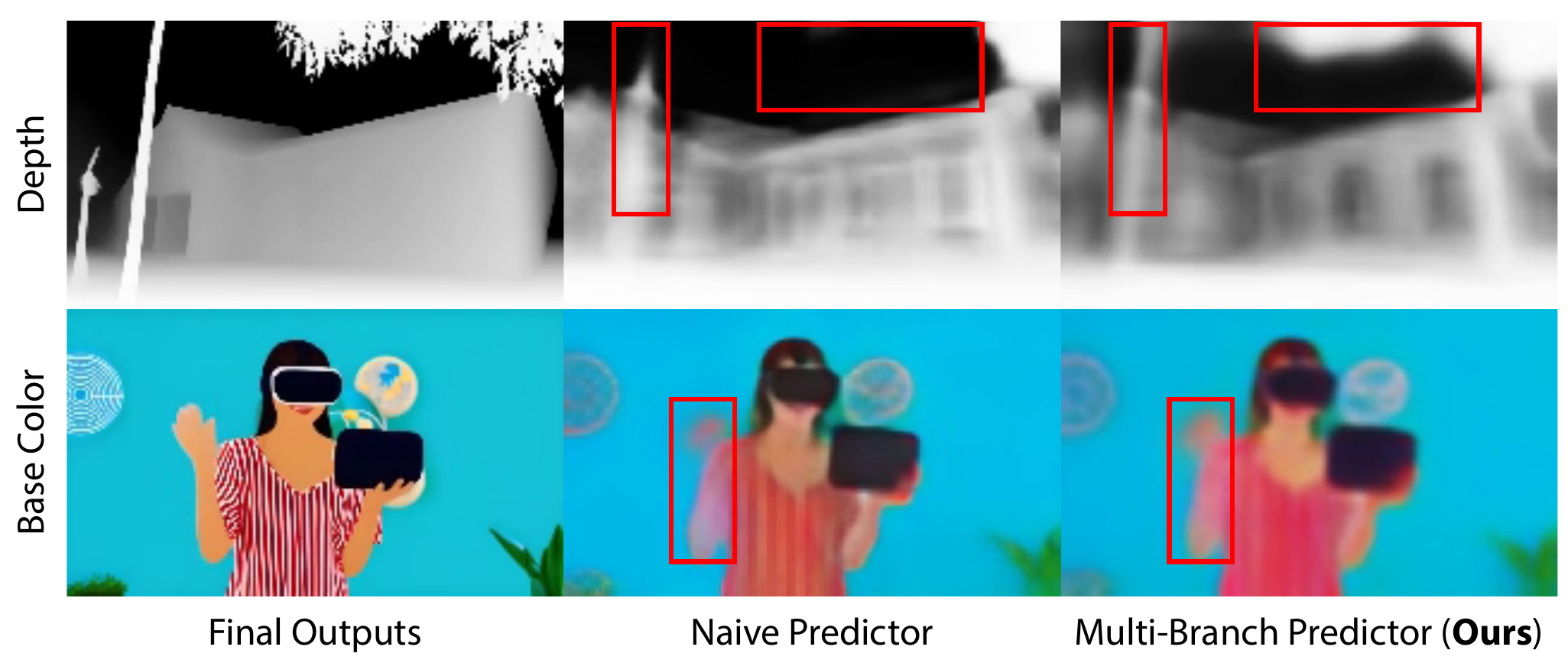}
\vspace{-15pt}
\caption{\textbf{Qualitatively, the proposed MB decoder improves mode selection and reduces artifacts due to multimodal ambiguity.} Red boxes highlight high-uncertainty regions that caused blurred patches in the naive single-branch decoder.}
\vspace{-10pt}
\label{fig:single_branch_multi_branch_comparison}
\end{figure}

Using the ensemble prediction from a finite number of branches during inference, the multi-branch model achieves higher-quality results and clearer edges while also encouraging diversity across branches, compared to the naive predictor, which tends to produce sharp edges but misaligns with the final video output, as shown in Figure~\ref{fig:single_branch_multi_branch_comparison}.

\section{Generating Variations Using Previews}

\paragraph{Multi-step diffusion as traversing a tree.}
Our multi-branch decoder can generate multi-channel video previews at any timestep in less than $1$ second of wall-clock time that are consistent with the $4$-second final videos (see Table~\ref{tab:model_comparison}). These properties make it practical for users to interact with the generation system in a novel tree structure (see Figure~\ref{fig:teaser}) where they can move up/down the denoising levels. To enrich this, we propose two variation generation methods to steer between siblings within the same noise level.

\subsection{Variations Through Stochastic Renoising}
\label{sec:variations_renoise}
The first way to introduce variation is simply by renoising a clean latent prediction $\hat{\mathbf{z}}_0$ using different random noise:
\begin{align}
\tilde{\mathbf{z}}_t = (1 - \sigma_{t_p}) \hat{\mathbf{z}}_0 + \sigma_{t_p} \boldsymbol{\epsilon}, \quad \boldsymbol{\epsilon} \sim \mathcal{N}(0, \mathbf{I}).
\end{align}
Here, $t_p$ is the timestep immediately after previewing occurs. The noise scale $\sigma_{t_p}$ is consistent with the original schedule, preserving the image structure generated so far while introducing stochastic finer-scale variations. Multiple denoised samples from the perturbed latents yield a set of plausible scene variations without repeating the full diffusion process.

\subsection{Variation Through Latent Steering}
\label{sec:variations_steer}
We also introduce an incremental method to \emph{steer} the later denoising timesteps using the trained preview decoder, generating purposeful variations. Specifically, with a trained, frozen preview decoder, steering involves solving the optimization problem
\begin{align}
\min_{\mathbf{f}_{t,b}} \mathcal{L}\big(\mathcal{D}(\mathbf{f}_{t,b}), \mathbf{y}^\ast\big),
\end{align}
where $\mathcal{D}: \mathcal{F} \to \mathcal{P}$ maps features $\mathbf{f}_{t,b} \in \mathcal{F}$ to the intrinsic map space $\mathcal{P}$. The details of the optimization can be found in the supplementary material~\S\ref{sec:supp_more_details}.
\begin{figure*}[!ht]
\centering
\includegraphics[width=0.95\linewidth]{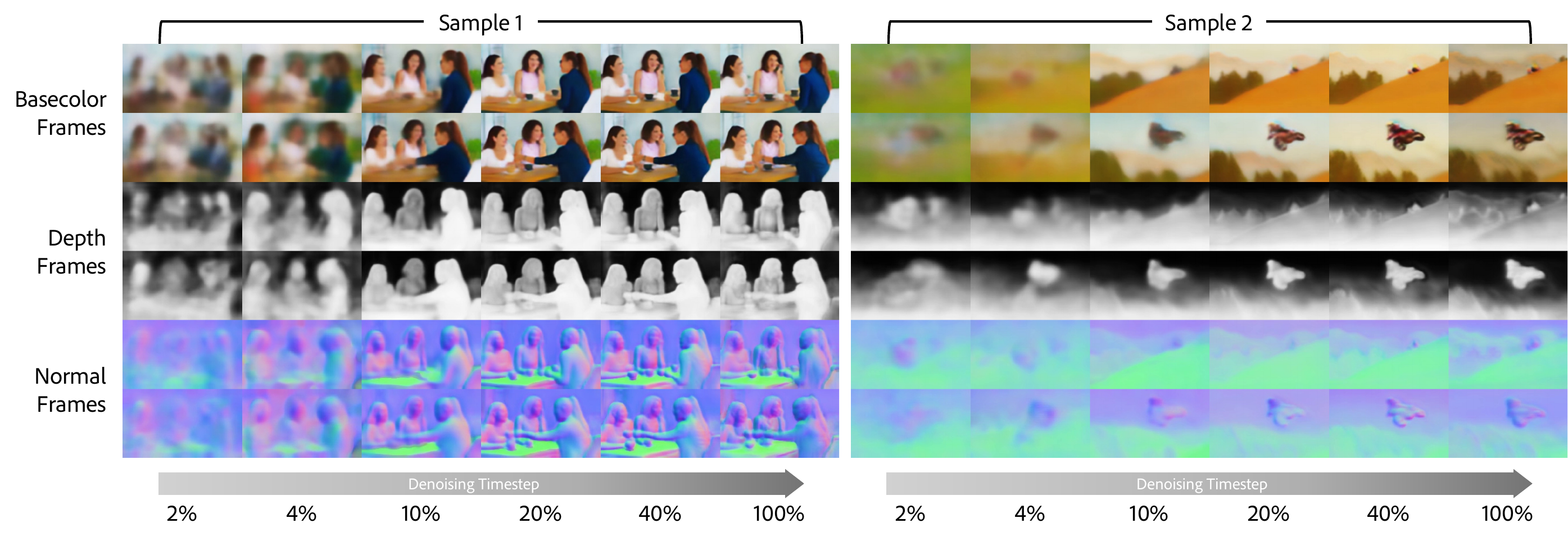}
\vspace{-10pt}
\caption{Timestep-wise evolution of base color, normal, and albedo. Coarse geometry and recognizable structures appear around the 5th timestep, with details refined progressively thereafter.}
\vspace{-10pt}
\label{fig:timesteps}
\end{figure*}

\section{Experiments}
\label{sec:experiments}

We conduct extensive experiments to validate our hypotheses and demonstrate the effectiveness of the proposed framework.

\subsection{Implementation Details}
\label{sec:implementation_details}
In this paper, we use Wan 2.1~\cite{wan2025}, although our framework is model-agnostic. We constructed a synthetic video dataset with cached intermediate diffusion features. A total of 1,000 videos were generated with unique prompts using DiffusionRenderer~\citep{liang2025diffusion}, which provided scene intrinsic channels along with RGB pixels. Our MB decoder is implemented with four 3D convolutional layers followed by two upscaling 3D convolutional layers for each branch, with $K=4$ branches. This results in a resolution of roughly $208\times120$; therefore, we use linear interpolation to downsample RGB and pseudo-ground-truth. Temporally, we subsampled every fourth frame to match the temporal size of the features. Ensemble weighting $\lambda_{\text{ens.}}=10.0$.

\begin{table*}[!ht]
\centering
\footnotesize
\begin{tabular}{lccccccrr}
\toprule
\textbf{Decoder Type} & \textbf{RGB} & \textbf{Base Color} & \textbf{Depth} & \textbf{Normal} & \textbf{Metallicity} & \textbf{Roughness} & \textbf{Runtime} & \textbf{Speedup} \\
\hline
\xzeropred & 16.98 & -- & -- & -- & -- & -- & 4.69s & 8.85x \\
Video Depth Anything* & -- & -- & 11.64 & -- & -- & -- & 9.50s & 17.92x \\
Diffusion Renderer* & -- & 14.81 & 5.17 & 18.45 & \textbf{17.22} & 14.72 & 222.87s & 420.51x \\
Linear Probing & 17.96 & 15.51 & 15.52 & 19.54 & 11.22 & 15.75 & 0.47s & 0.89x\\
\textbf{Ours} & \textbf{18.03} & \textbf{16.38} & \textbf{16.95} & \textbf{20.04} & 16.42 & \textbf{17.03} & 0.53s & 1x\\
\bottomrule
\end{tabular}
\vspace{-5pt}
\caption{\textbf{Comparison across different models and modalities} using PSNR and wall-clock time at 10\% of the total denoising steps shows that our model produces the best results for most of the channels while being significantly resource-efficient. Speedup is computed using our approach as the baseline.}
\vspace{-10pt}
\label{tab:model_comparison}
\end{table*}

\subsection{Baseline Comparison}
We compare MB-based preview generation with other methods. Since our model is uniquely multi-modal, we compare channels with separate baselines when possible; these include $\mathbf{x}_0$ prediction (``\xzeropred") that uses Tweedie's formula to estimate the clean latent and then passes it through the pretrained VAE decoder to obtain a clean RGB video. We also compare to the state-of-the-art Video Depth Anything~\citep{chen2025videodepthanything} for depth estimation and intrinsics with DiffusionRenderer~\citep{liang2025diffusion}, both using \xzeropred{} as input. We selected $10\%$ of the total denoising steps and report PSNR as the metric. The results are shown in Table~\ref{tab:model_comparison}. Our predictor outperforms the other baselines, suggesting the effectiveness of our feature-level predictor. Also, we measured the overhead using wall-clock time, which shows that our decoder is significantly more efficient compared to the baselines.

\begin{figure*}[!ht]
\centering
\includegraphics[width=0.95\linewidth]{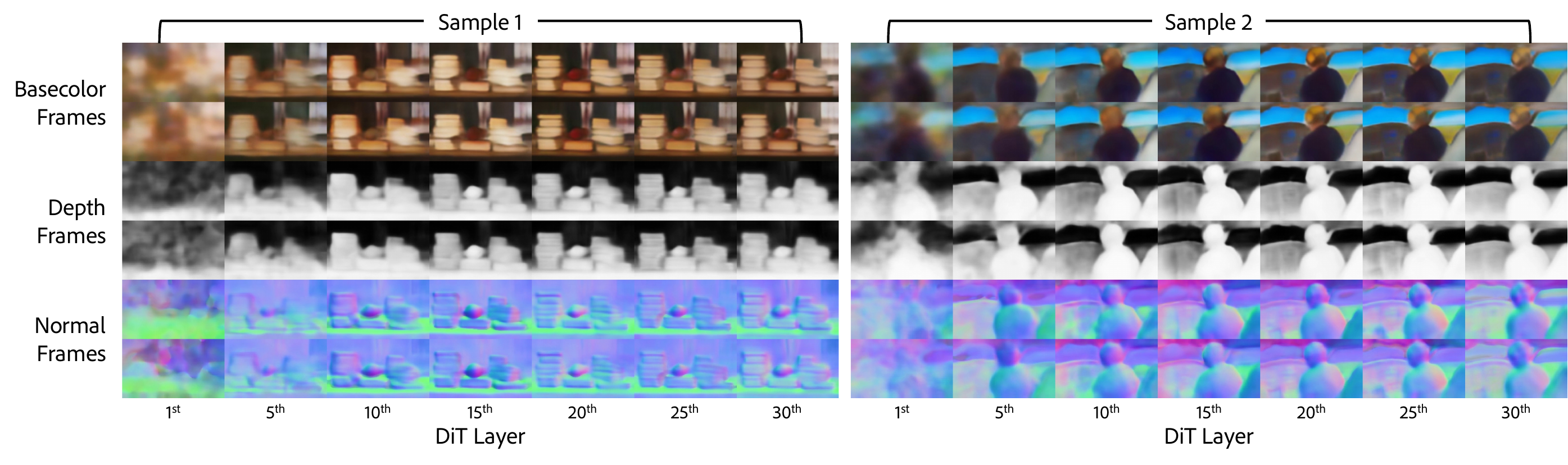}
\vspace{-10pt}
\caption{Block-wise evolution of base color, normal, and albedo. Intrinsics are best predicted from mid-level features, slightly degrading in the final layers.}
\vspace{-10pt}
\label{fig:block-evolution}
\end{figure*}

\subsection{Stepwise and Blockwise Preview Evolution}
Representative examples of how previews evolve stepwise and blockwise are shown in Figure~\ref{fig:block-evolution}. We found consistent convergence behaviors as shown by the linear probing analysis in Figure~\ref{fig:linear-probing}. In addition, qualitatively, rough geometry and scene structure appear as early as $\sim10\%$ of the denoising steps, which are well captured by the scene intrinsics. These material previews remain stable throughout the denoising process, consistent with the final generation.

Figure~\ref{fig:timesteps} shows the blockwise evolution. We found that lower blocks contain coarse geometry and color distributions, while mid-level blocks (around the 15\textsuperscript{th}--20\textsuperscript{th} of the 30-layer model) capture detailed spatial structure with stable base color and depth predictions. At the last block, the predictive power for intrinsic properties like depth and normals decreases. These observations also align with the linear probing analysis in Figure~\ref{fig:linear-probing}, showing that intrinsic information saturates in mid-to-late blocks, after which the representation primarily supports appearance refinement.

\begin{figure}[t]
\centering
\includegraphics[width=0.8\linewidth]{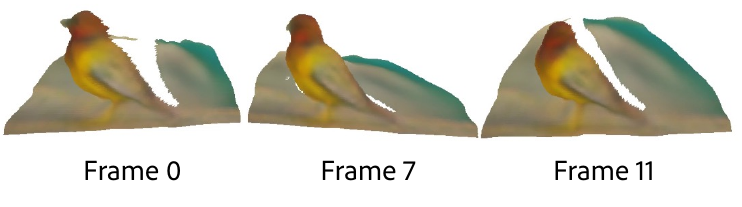}
\vspace{-10pt}
\caption{\textbf{Rubber-like 4D visualization} can be derived from the intrinsic previews from our model. Interestingly, at only 10\% of the denoising schedule, a clear structural representation of the scene has already emerged.
}
\vspace{-5pt}
\label{fig:rubber}
\end{figure}

\subsection{Rubber-Like 4D Previsualization}
We show that the MB decoder preview at only 10\% of the denoising steps can be used to create a 4D visualization of the video being generated (Figure~\ref{fig:rubber}). Despite being computed from highly noisy intermediate features, the previews reveal smooth object motion, spatial composition, and overall color palette, resembling a ``rubber-like" low-frequency representation of the scene, which can be useful for interactive exploration. More results can be seen in the supplementary material~\S\ref{sec:supp_more_analysis}.

\subsection{Variation Generation}

\begin{figure}[!ht]
\centering
\includegraphics[width=1.0\linewidth]{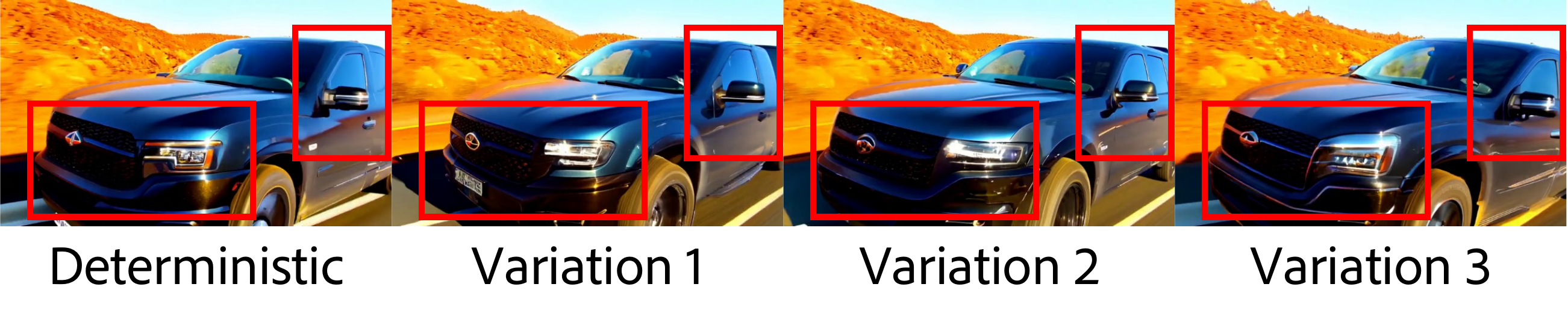}
\vspace{-20pt}
\caption{\textbf{Examples of variation generation via stochasticity injection} show coarse details being preserved at lower noise levels, while the injected stochasticity changes several details in the video highlighted by the red boxes.
}
\vspace{-10pt}
\label{fig:stochastic}
\end{figure}

\paragraph{Stochastic variation generation} introduced in~\S\ref{sec:variations_renoise} renoises latents using the appropriate scale based on noise level. At intermediate steps, users can preview the coarse scene structures using our MB decoder and then experiment with alternative finer details by sampling the base model at that level. An example is shown in Figure~\ref{fig:stochastic}.

\begin{figure*}[t]
\centering
\includegraphics[width=0.9\linewidth]{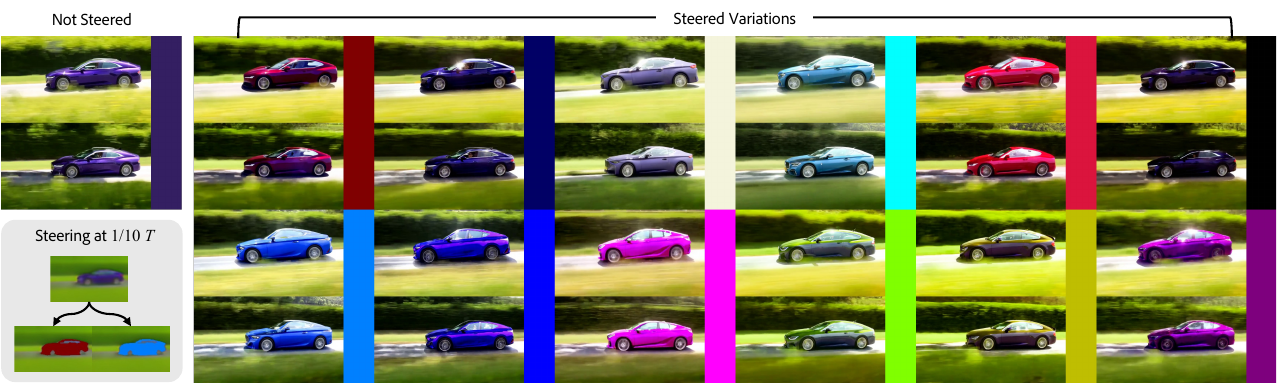}
\vspace{-10pt}
\caption{\textbf{Steering base color} at 10\% of the total denoising steps allows users to generate variations in the same context. The text prompt is ``A car driving on a sunny road".
}
\vspace{-5pt}
\label{fig:steering-car}
\end{figure*}

\vspace{-10pt}
\paragraph{Steered variation generation} introduced in~\S\ref{sec:variations_steer} allows for channel-targeted steering. Figures~\ref{fig:steering-car} and~\ref{fig:steering} show separate color and geometry steering, and more results are provided in the supplementary material~\S\ref{sec:supp_more_details}. Note that lighting and texture remain consistent across steered variants, likely because they are handled by later stages of the denoising process. Also note that steering is a different task from video editing. The latter comprises myriad different techniques and is aimed at changing the final output with precision, whereas the steering proposed in this work is a meaningful way of generating variations \emph{during} generation and is meant to be complementary to video editing methods (e.g., steer-then-edit or preview-while-editing).

\begin{figure}[!ht]
\centering
\includegraphics[width=1.0\linewidth]{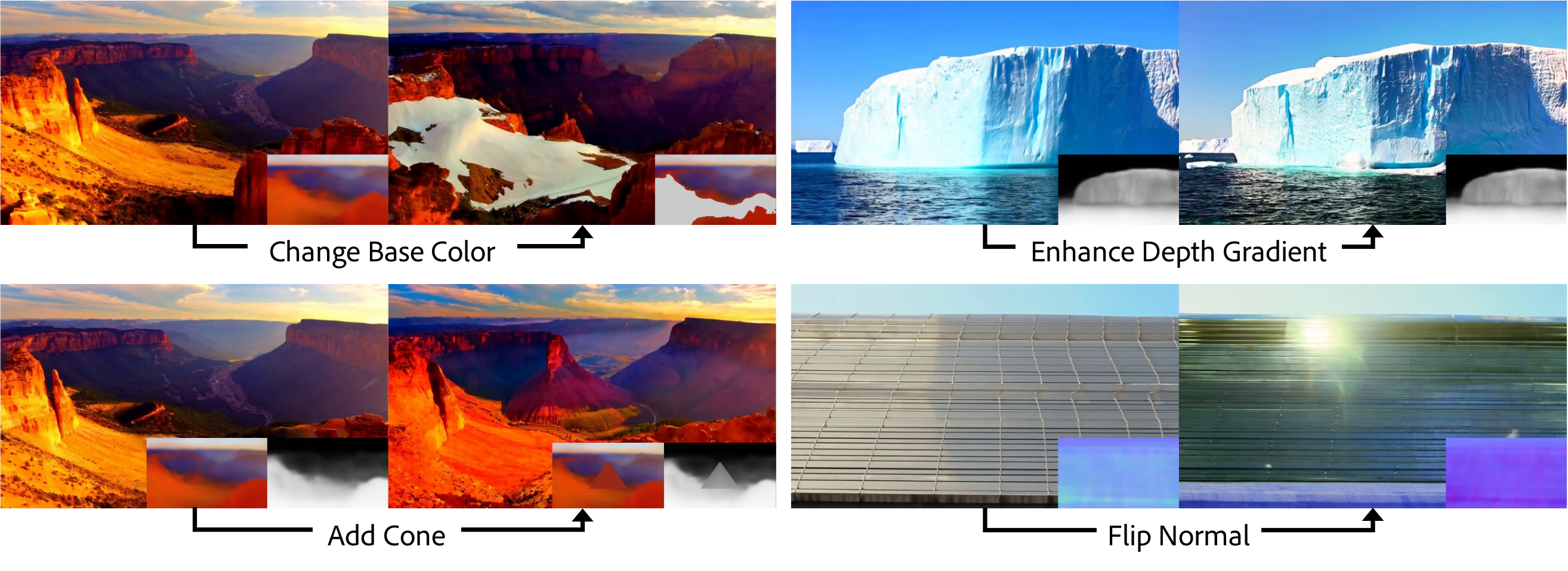}
\vspace{-20pt}
\caption{\textbf{Examples of variation generation via steering} show meaningful steered base color, depth, and normal results.
}
\vspace{-10pt}
\label{fig:steering}
\end{figure}

\begin{table}[!ht]
\scriptsize
\centering
\begin{tabular}{lccccc}
\toprule
\textbf{Model}  &  \textbf{Albedo} & \textbf{Depth} & \textbf{Normal} & \textbf{Metallicity} & \textbf{Roughness} \\
\midrule
Naive   & 0.121 & 0.121 & 0.167 & 0.249 & 0.152 \\
\midrule
Shallow   & 0.120 & 0.115 & 0.156 & \textbf{0.238} & \textbf{0.140} \\
Deep   & 0.121 & 0.118 & 0.160 & 0.240 & 0.146 \\
\textbf{Ours}   & \textbf{0.117} & \textbf{0.115} & \textbf{0.156} & 0.241 & 0.142 \\
\bottomrule
\end{tabular}
\vspace{-5pt}
\caption{
\textbf{Ablation study} comparing decoder variants. We report $L_1$ error on the validation set. Both the naive and our (MB) decoders are 6 layers deep; the shallow is 4 layers, and the deep is 8.
}
\vspace{-10pt}
\label{tab:ablation-comparison}
\end{table}

\subsection{Ablation Study}
We analyze the impact of key architectural choices in Table~\ref{tab:ablation-comparison}. The MB decoder achieves lower MSE and $L_1$ errors across most intrinsic properties compared to a single-branch counterpart, confirming that modeling multiple hypotheses improves robustness and interpretability (Figure~\ref{fig:single_branch_multi_branch_comparison}). Shallower or deeper variants show marginal differences, suggesting that our six-layer multi-branch configuration provides a balanced trade-off between accuracy and computational cost. 

\subsection{User Study}
To evaluate the perceptual quality of our representations, we conducted a user study with 35 participants comparing our method against the \xzeropred{} baseline. Participants were shown two representations alongside a reference video and asked to judge which better predicted video content, exhibited fewer visual artifacts, and more clearly conveyed scene composition. Previews generated by \projname{} were preferred $74.6\%$, $72.9\%$, and $76.9\%$ of the time for content predictability, visual fidelity, and scene clarity, respectively, when compared to the \xzeropred{} baseline. More details can be found in the supplementary material~\S\ref{sec:user_studies}.

\section{Discussion}
\label{sec:discussion}
\paragraph{Utilizing Diffusion Features}
Video diffusion models must resolve greater ambiguity than image diffusion models and are expected to learn more informative features. However, their learned representations remain relatively underexplored. To the best of our knowledge, our work is the first to utilize video diffusion transformer features to predict multiple scene intrinsics simultaneously, providing a rich analysis of diffusion features. In video diffusion, these features correlate strongly with physical scene attributes such as depth and albedo, supporting the hypothesis that diffusion implicitly performs a form of inverse rendering.
\vspace{-5pt}
\paragraph{Superposition Problem}
The superposition problem was introduced and empirically verified in~\S\ref{sec:mb_predictor}, which describes the observation of blurred predictions at intermediate denoising states for high-motion and high-complexity regions. We attributed this to diffusion features encoding multiple possible future states simultaneously, and reason that it is a superset of the notorious hallucination problem such as the implausible 6-finger hand generated by a diffusion model. We provided theoretical reasoning for the source of this problem and empirically verified it via a small-scale toy problem. We showed that an explicit, multi-headed architecture like our MB decoder can mitigate this issue. We believe there is potential to extend this approach to related problems, such as few-step distillation, where models are also constrained by limited NFEs and can lead to exacerbated hallucination and quality degradation. We leave this as an exciting direction for future work. 
\begin{figure}[t]
\centering
\includegraphics[width=1.0\linewidth]{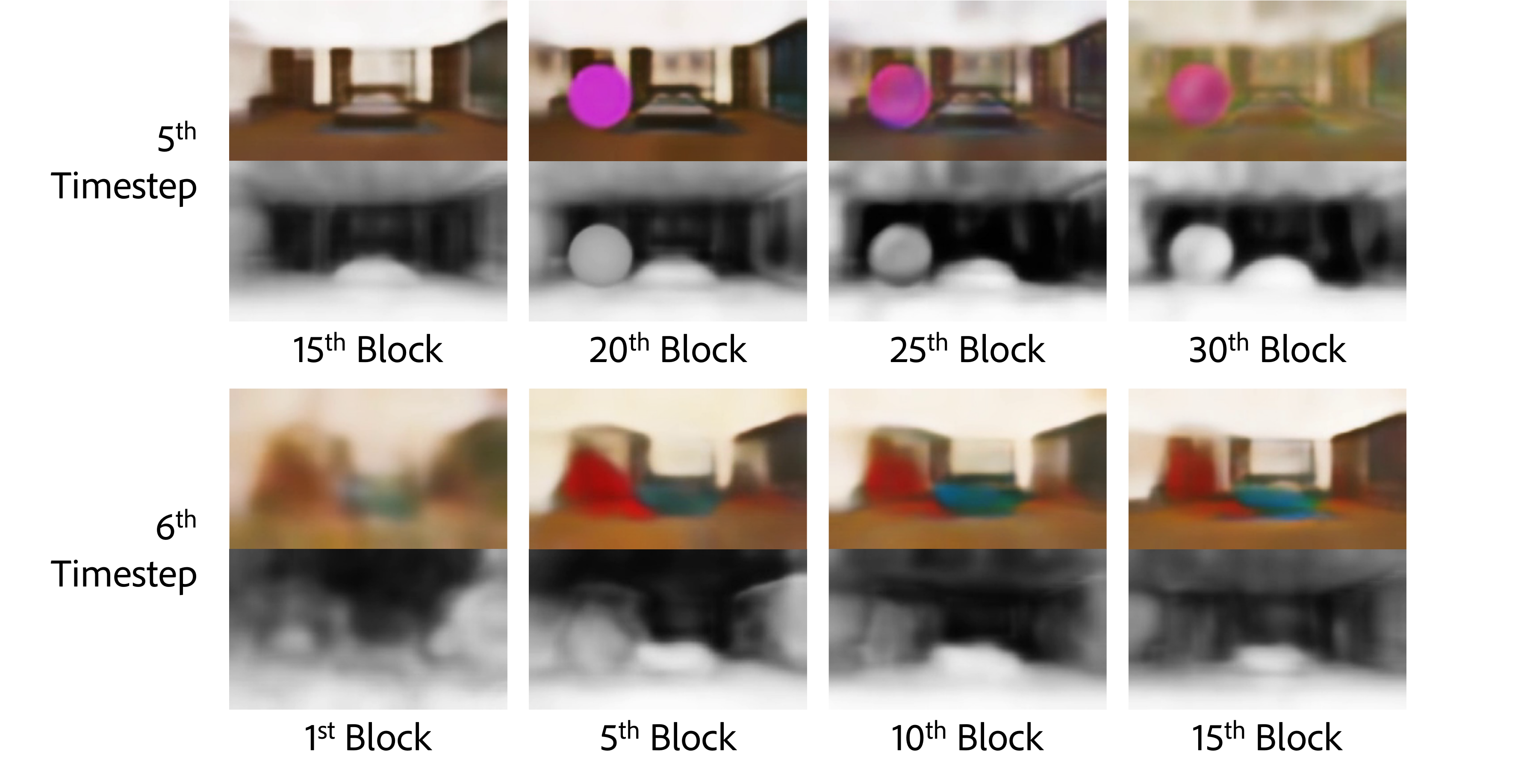}
\vspace{-20pt}
\caption{Failure case in intrinsic steering. The sphere added at the 20th layer gradually dissolves and deforms in subsequent timesteps.}
\vspace{-10pt}
\label{fig:steering-failure}
\end{figure}
\vspace{-5pt}
\paragraph{Limitations and Future Work}
While our framework enables fast and semantically meaningful previews for video diffusion models, several limitations remain. We deliberately limit our scope to scene intrinsics, and text prompts are not considered; the interaction between intrinsic previews and text-driven conditioning can be explored in future work. Additionally, there are failure cases in steering, as shown in Figure~\ref{fig:steering-failure}, where the steered intrinsics dissipate as denoising progresses, which we attribute to an out-of-distribution issue for our shallow decoder. For future work, we aim to explore alternative decoder architectures to improve mode separation and produce clearer, more coherent outputs at higher resolution, as well as expand the intrinsic representations to include additional modalities.
\vspace{-5pt}
\section{Conclusion}
\projname{} offers a new perspective on interacting with video diffusion models by making their coarse-to-fine internal evolution visible, actionable, and efficient. By decoding stable intrinsic signals that emerge early in the denoising process, our lightweight, plug-and-play preview framework enables users to terminate unpromising generations, iterate rapidly, and steer trajectories without sacrificing final quality. Beyond practical speedups, these previews serve as a window into the geometry, layout, and appearance dynamics that govern diffusion behavior, opening new opportunities for interpretability and user-driven control. We believe \projname{} lays the groundwork for more interactive, transparent, and resource-efficient video generation pipelines, and provides a foundation for future research into controllable diffusion and the structure of generative processes.
{
    \small
    \bibliographystyle{ieeenat_fullname}
    \bibliography{main}
}

\clearpage
\setcounter{section}{0}
\renewcommand{\thesection}{S\arabic{section}}

\twocolumn[
\begin{center}
{\Large \textbf{DiffusionBrowser: Interactive Diffusion Previews via Multi-Branch Decoders}}\\\vspace{5pt}
{\large Supplementary Material}\vspace{10pt}
\end{center}
]

\section{Analysis: Diffusion Features and Intrinsic Previews}
\label{sec:supp_more_analysis}

\begin{figure}[t]
\centering
\includegraphics[width=1.0\linewidth]{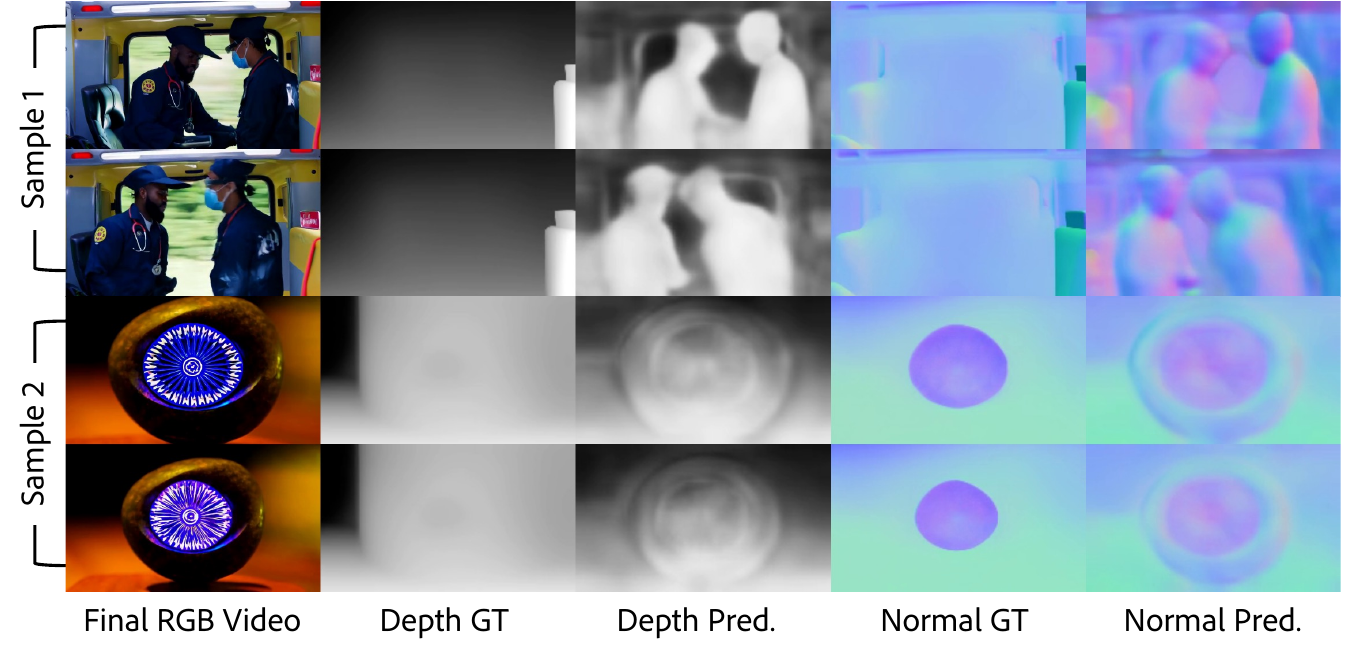}
\caption{Even when pseudo-GT data predicted with DiffusionRenderer~\cite{liang2025diffusion} contains incorrect geometry, our decoder predicts plausible and consistent structure from diffusion features.}
\label{fig:geometry-refine}
\end{figure}

In this section, we provide further evidence that intermediate diffusion features contain cleaner intrinsic structure than the pseudo-ground-truth supervision. Although the pseudo-ground-truth intrinsics used for training may contain geometric inaccuracies, Figure~\ref{fig:geometry-refine} shows that our decoder can produce more stable and coherent geometry than the supervision itself. This highlights that diffusion features encode intrinsic scene structure reliably using their strong prior and that our decoder manages to extract this information even when training labels are imperfect.

\begin{figure*}[t]
\centering
\includegraphics[width=1.0\linewidth]{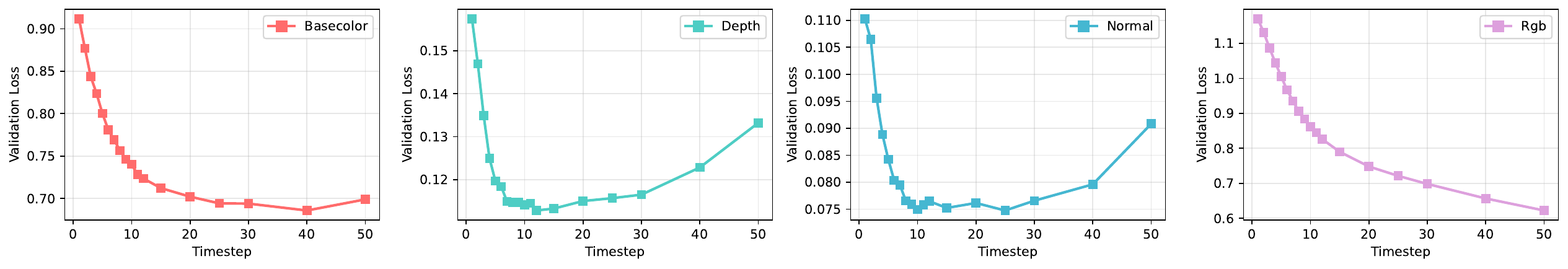}
\includegraphics[width=1.0\linewidth]{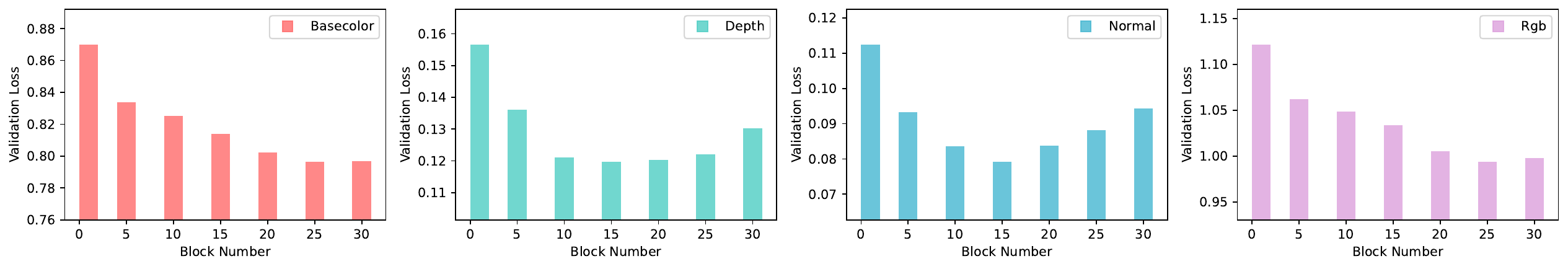}
\caption{Nonlinear probing comparisons across timesteps and blocks. The reported loss is the last-epoch validation loss, $\ell_1$ + perceptual. The results show a similar trend to linear decoding.}
\label{fig:toy-ens}
\end{figure*}

To further quantify the decodability of intrinsic signals, we retrain a series of nonlinear decoders across all blocks and timesteps, shown in Figure~\ref{fig:toy-ens}. These experiments extend the linear-probing analysis from the main paper (Figure~\ref{fig:linear-probing}) and confirm that predictive power saturates early across the network hierarchy. While deeper decoders improve performance relative to linear predictors, the overall trend remains the same; the most reliably decodable structure appears in early or mid-level layers.

\begin{figure*}[t]
\centering
\includegraphics[width=1.0\linewidth]{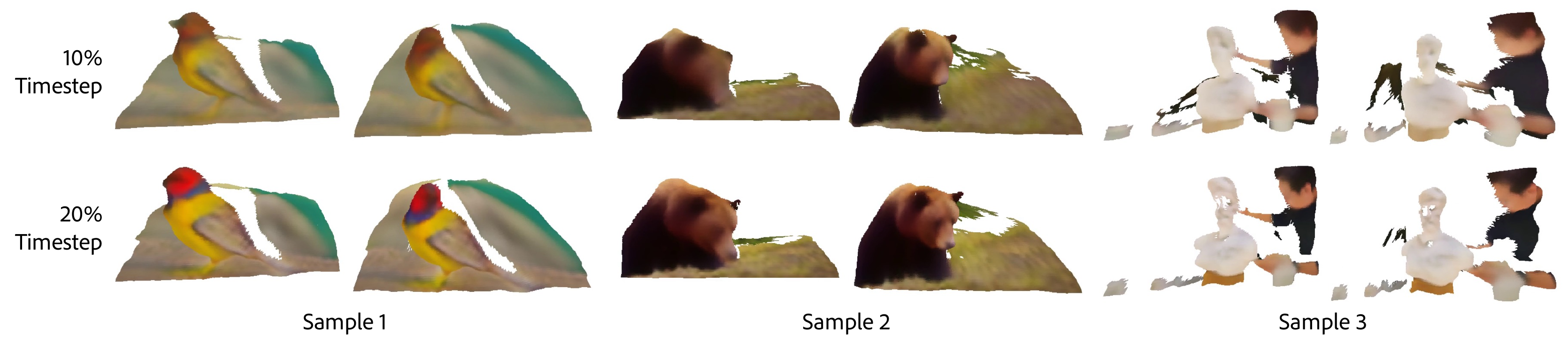}
\caption{Rubber-like 4D reconstruction results. Even at 10\% of the timestep, each reconstruction represents the composition, geometry, and dynamics of the scene, while at the 20\% timestep, a refined reconstruction result is produced.}
\label{fig:rubber-sup}
\end{figure*}

\begin{figure}[t]
\centering
\includegraphics[width=0.85\linewidth]{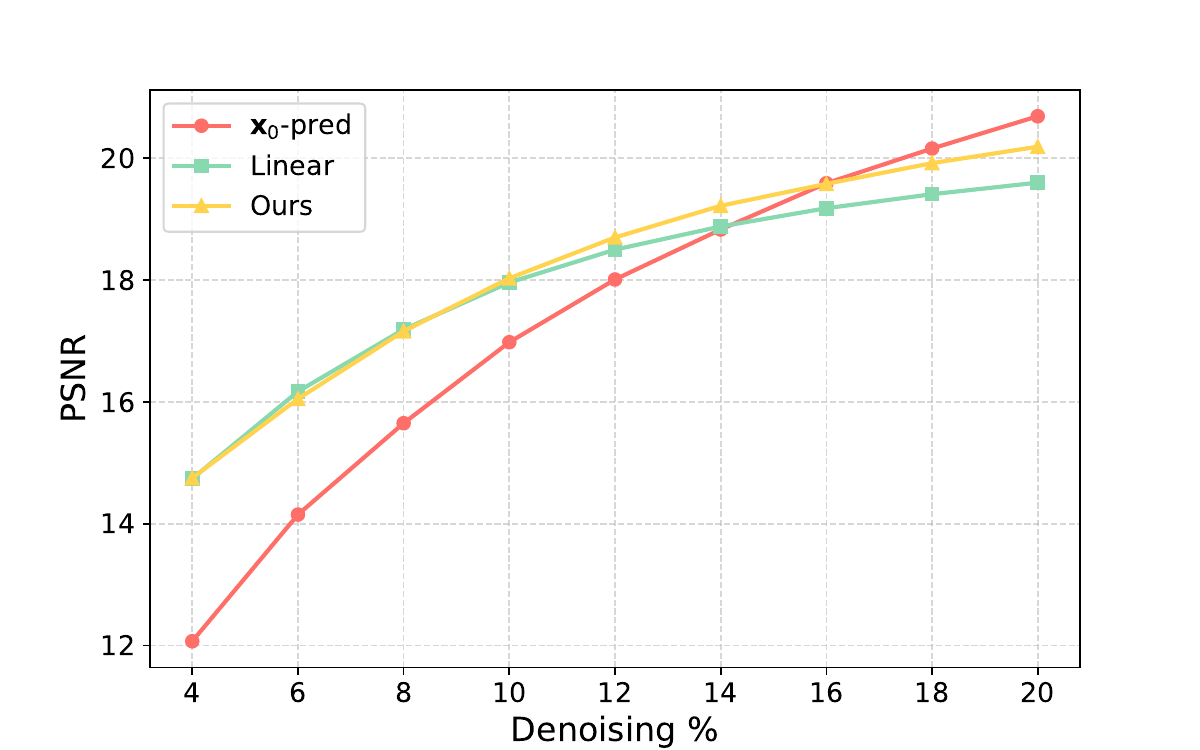}
\caption{PSNR comparison between $\mathbf{x}_0$-pred (the VAE decoder), Linear, and our method. In the high-noise regime, the linear and our decoders perform similarly, with the gap increasing as the denoising process progresses. The PSNR of the $\mathbf{x}_0$-pred decoder and our method crosses at 16\% of the denoising steps, suggesting that early previews benefit substantially from our decoder.}
\label{fig:rgb-psnr}
\end{figure}

Finally, Tables~\ref{tab:psnr-decoder}--\ref{tab:lpips-decoder} and Figure~\ref{fig:rgb-psnr} compare decoder performance across timesteps. The multi-branch architecture becomes increasingly beneficial as noise decreases, widening the gap over linear predictors. Importantly, $\mathbf{x}_0$-prediction performs significantly worse than both feature decoders at early timesteps, even for RGB, and only surpasses them at approximately 16\% of the denoising process. Although this is a relatively early stage of denoising, it reveals much of the geometry of the dynamic scene and allows us to reconstruct rubber-like results (Figure~\ref{fig:rubber-sup}). This supports our claim that early previews substantially benefit from decoding features rather than relying on the VAE decoder.

\section{Synthetic Training Data Generation}
\label{sec:supp_data_details}
To train the intrinsic decoders, we constructed a dataset designed to cover a broad range of scene types. Table~\ref{tab:categories} lists the 40 scene categories used for prompt generation, spanning human activities, animals, natural environments, indoor and outdoor scenes, motion types, fantasy concepts, and more. For each category, we generated 25 prompts, yielding 1{,}000 prompts in total. We then ran DiffusionRenderer~\cite{liang2025diffusion} to predict all intrinsics for decoder training, as described in \S\ref{sec:implementation_details}.

The proposed dataset exposes the decoder to the diversity of structures and appearance variations encountered during video diffusion. Training details for the decoders are provided in the main paper.

\section{Analysis: More on the Toy Problem}
\label{sec:supp_toy_problem}
In the toy experiment demonstrating the superposition phenomenon in the main paper (\S\ref{sec:toy_problem}), we leverage a controlled toy environment consisting of 4-frame sequences at $7 \times 7$ resolution for each frame. A single white dot moves left, right, or remains stationary. We examine two variants: (1) motion-only uncertainty and (2) motion+position uncertainty, where the starting location is slightly jittered to induce multimodal clean states.

\begin{figure}[t]
\centering
\includegraphics[width=1.0\linewidth]{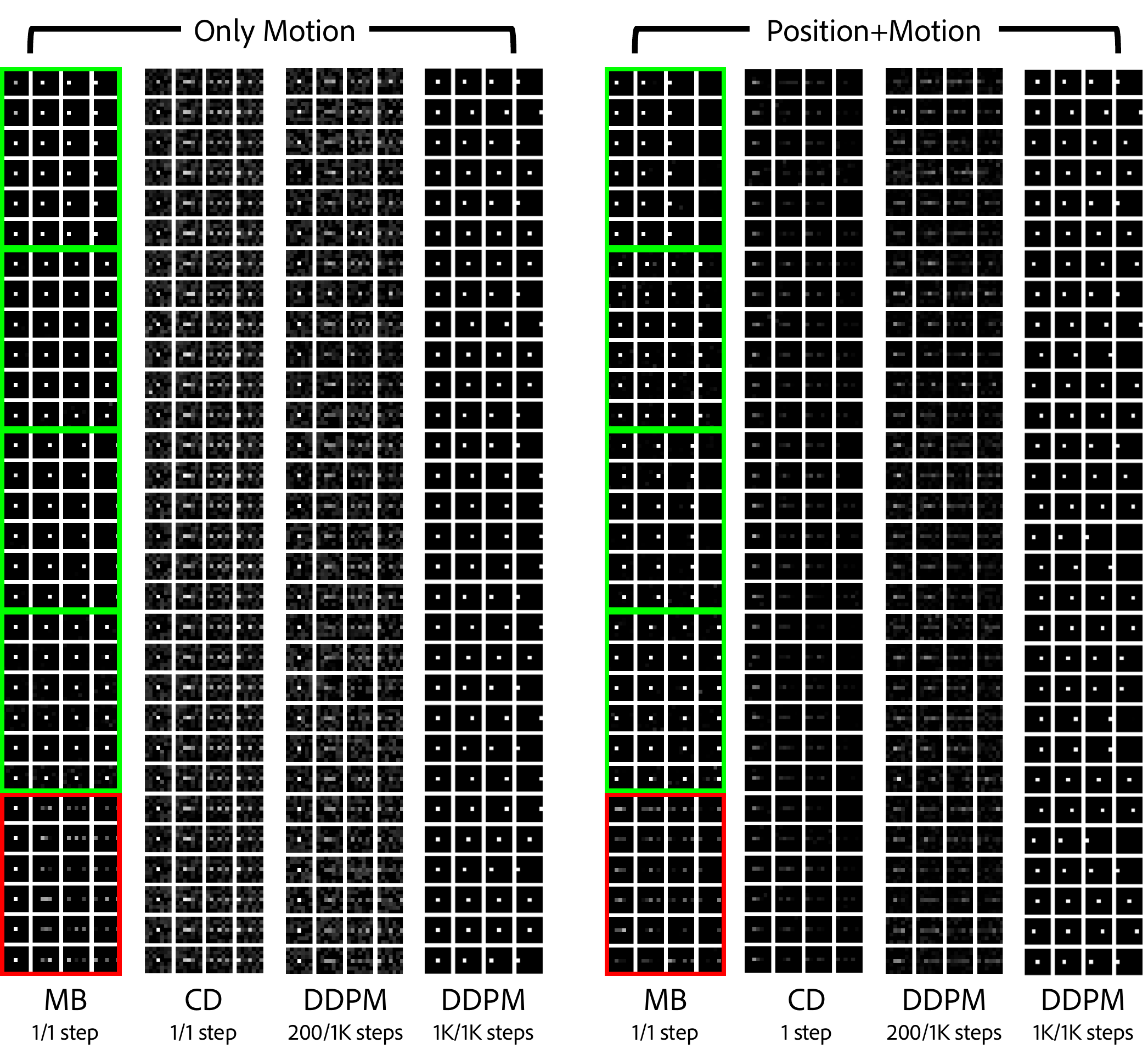}
\caption{Toy experiment. Multi-branch predictions recover separate modes without superposition. The green boxes represent the prediction of each branch, and the red box represents the averaged prediction of the branches.}
\label{fig:toy-sup}
\end{figure}

This controlled setting reveals how diffusion models trained with MSE behave when the clean posterior is multimodal: at high-noise timesteps, the model predicts the posterior mean, producing in-between states that never occur in the data. Figure~\ref{fig:toy-sup} illustrates typical artifacts in this setting, e.g., duplicated or faded dots, when using a low number of function evaluations (NFE).

\begin{figure}[t]
\centering
\includegraphics[width=0.85\linewidth]{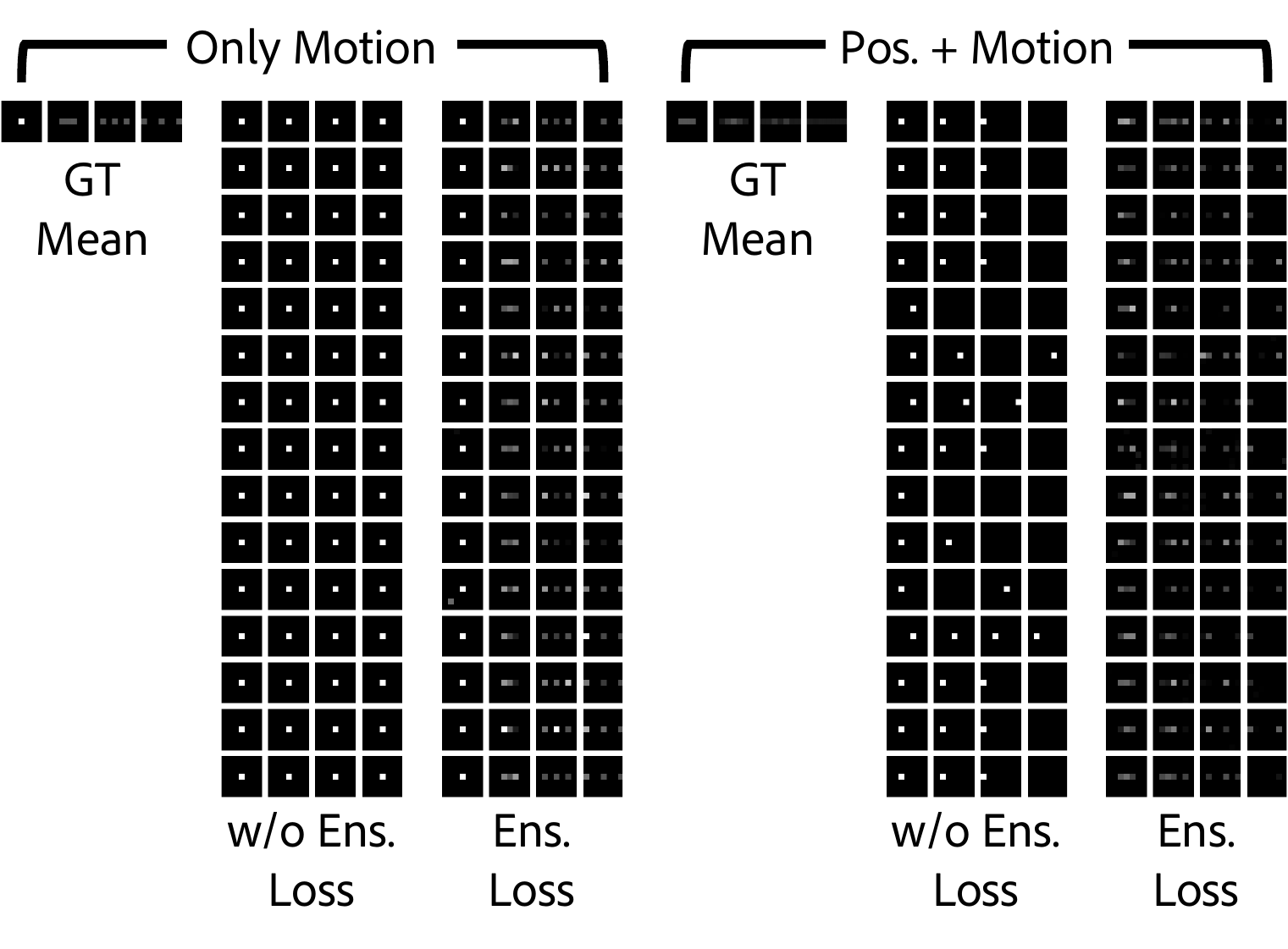}
\caption{Without ensemble loss, the reduced diversity results in collapse to a fewer number of modes and causes artifacts.}
\label{fig:toy-ens-2}
\end{figure}

For our toy multi-branch architecture, we use a simpler mode-seeking objective based on the available dataset. Because we have direct access to the full ground-truth distribution, we can explicitly encourage each branch to predict a specific data mode. This pushes each branch toward the data instead of toward the average, achieving a similar effect to $\ell_1$ + perceptual loss in our final decoder model. Therefore, the final training loss selects the closest data point in terms of $\ell_1$ distance (mode-seeking loss), while an ensemble term uses a standard MSE objective:
\begin{align}
\mathcal{L}_{\text{branch}}^{(k)} &= \min_{x_0 \sim \text{dataset}} \lVert x_0^{(k)} - x_0 \rVert_1, \\
\mathcal{L}_{\text{ens}} &= \frac{1}{K} \sum_{k=1}^{K} \lVert x_0^{(k)} - x_0 \rVert_2^2.
\end{align}
This structure encourages each branch to specialize in one plausible mode, yielding clean and mode-consistent predictions. The ensemble prediction helps each branch maintain diversity by regularizing their average toward the mean. When the ensemble loss is removed, as shown in Figure~\ref{fig:toy-ens-2}, the branches collapse toward fewer modes and fail to capture the accurate multimodal distribution.

Table~\ref{tab:toy} summarizes the numerical results. With 4 frames, the correct number of dots (of intensity exceeding 0.5) is 4 in expectation for motion-only sequences. The multi-branch decoder is the only method that consistently reproduces the correct mode structure under 1-NFE sampling. Distilled and few-step DDPM baselines produce artifacts under motion and position uncertainty, resulting in more dots. In contrast, the average number of boxes with motion and position uncertainty is approximately 3.8 (due to moving out of the frame at the last frame in two of nine cases), resulting in significantly fewer dots than expected. Our method alleviates this effect.

\begin{figure}[t]
\centering
\includegraphics[width=1.0\linewidth]{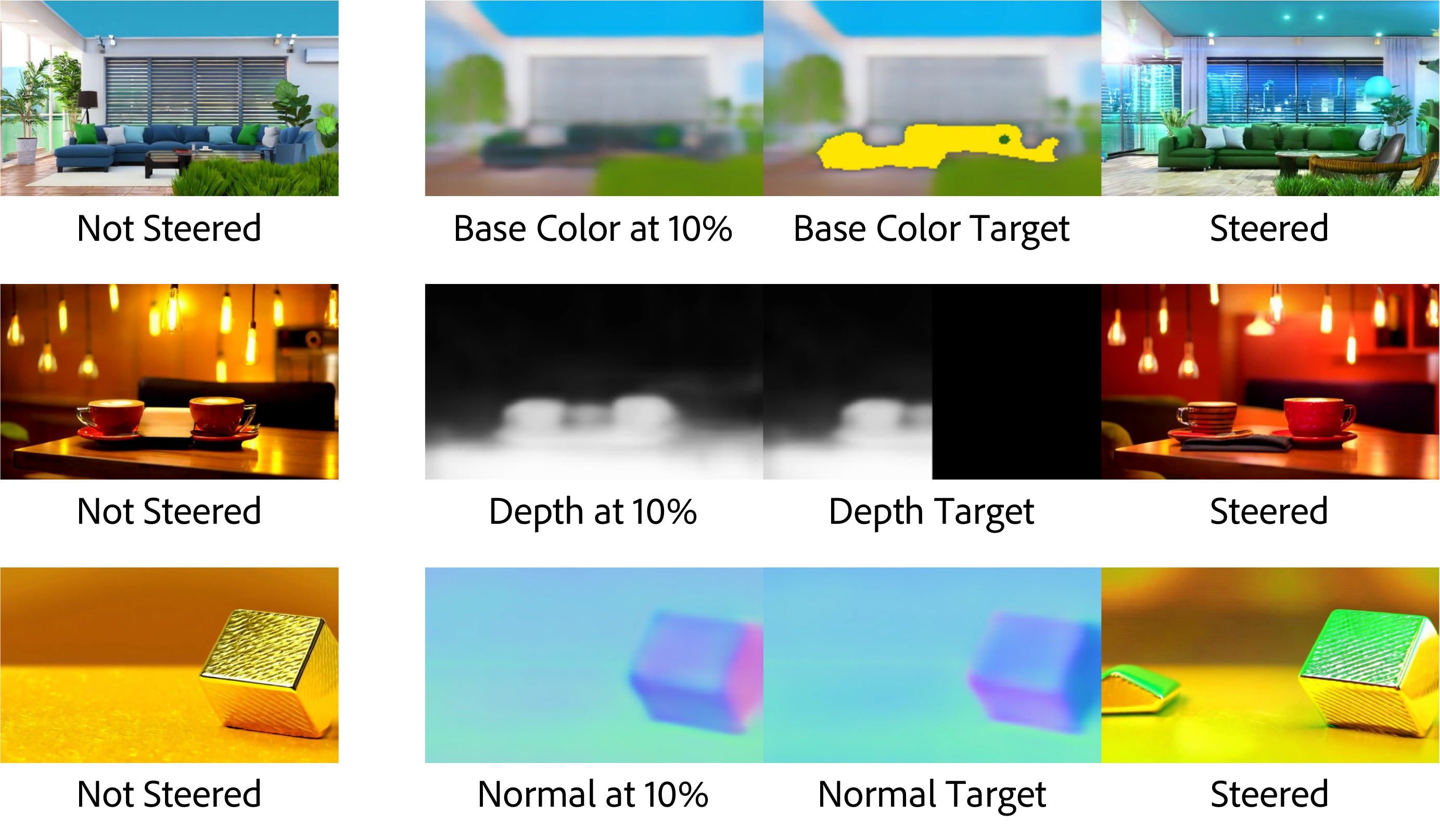}
\caption{Failure modes for different modalities. We steered each of the maps at 10\% of the denoising steps.}
\label{fig:steering-failure-mod}
\end{figure}

\begin{table}[t]
\small
\centering
\begin{tabular}{lcc}
\toprule
\textbf{Metric} & \textbf{\xzeropred} & \textbf{Ours} \\
\midrule
Content Predictability & 25.4\% & 74.6\% \\
Visual Fidelity & 27.1\% & 72.9\% \\
Scene Clarity & 23.1\% & 76.9\% \\
\midrule
\end{tabular}
\caption{User study comparing our intrinsic preview method against the $\mathbf{x}_0$-pred baseline.}
\label{tab:user_study}
\end{table}

\section{Details on Latent Steering}
\label{sec:supp_more_details}

We implement preview steering by applying small gradient-based modifications to intermediate diffusion features to guide the decoded intrinsic map toward a chosen target. A gradient update is applied in feature space using the Jacobian of $\mathcal{D}$, the learned multi-branch decoder. Normals are steered using a cosine loss, while other modalities use $\ell_1$ distance. Our goal is not to formalize a new optimization framework but to demonstrate that preview-level edits can be propagated back into diffusion features.

We explore simple proof-of-concept targets for different modalities:  
(1) \textbf{Base color:} cluster the predicted colors via K-means++ and shift toward another cluster.  
(2) \textbf{Depth edges:} enhance depth gradients using a Sobel operator.  
(3) \textbf{Normal flipping:} invert the $y$-axis of the predicted normal map. These are intentionally minimal examples; more sophisticated target-construction methods could be used. \emph{Complementary} to more traditional video editing methods, our latent steering method provides a simple yet efficient way of steering the denoising trajectory toward more favorable directions with minimal waste of compute.

We believe that steering during denoising presents a brand-new avenue for controllable generation. The results shown in our paper are far from perfect and have several failure modes. Precise color steering is not always possible, and dramatic geometric editing such as completely removing one half of the depth map or flipping normals to point the surface in a physically implausible orientation are some of the major failure cases, illustrated in Figure~\ref{fig:steering-failure-mod}. We attribute these to: 1) the small base model we used might not have sufficient 3D understanding capability; 2) limited capacity of the trained decoder; 3) out-of-distribution problems; and 4) simplistic steering methodology and imperfect execution. We leave the improvements and more careful examination of the steering results to future work.

\section{Benefits of Multi-Modal Previews}
\label{sec:supp_more_results}

\begin{figure*}[!ht]
\centering
\includegraphics[width=1.0\linewidth]{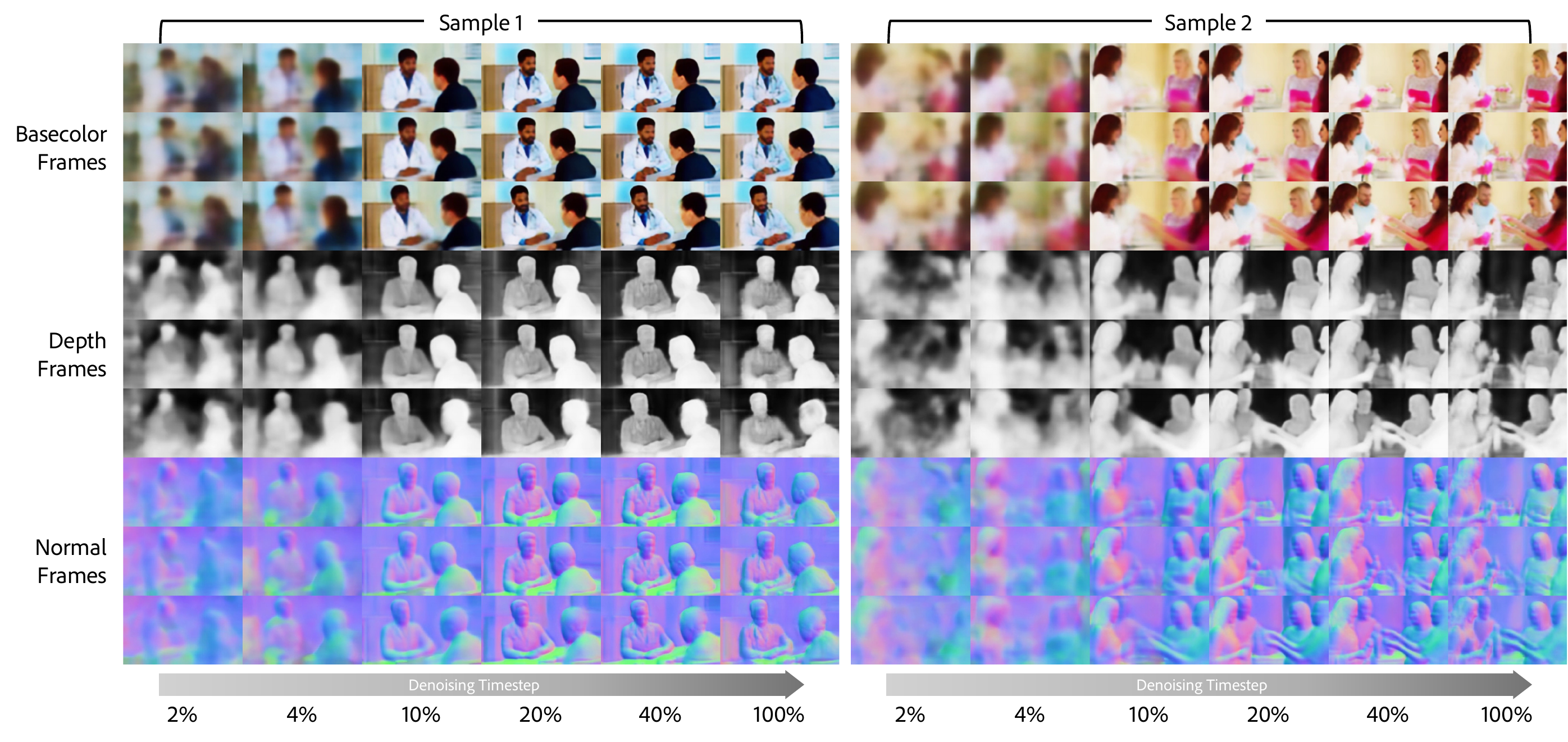}\\\vspace{20pt}
\includegraphics[width=1.0\linewidth]{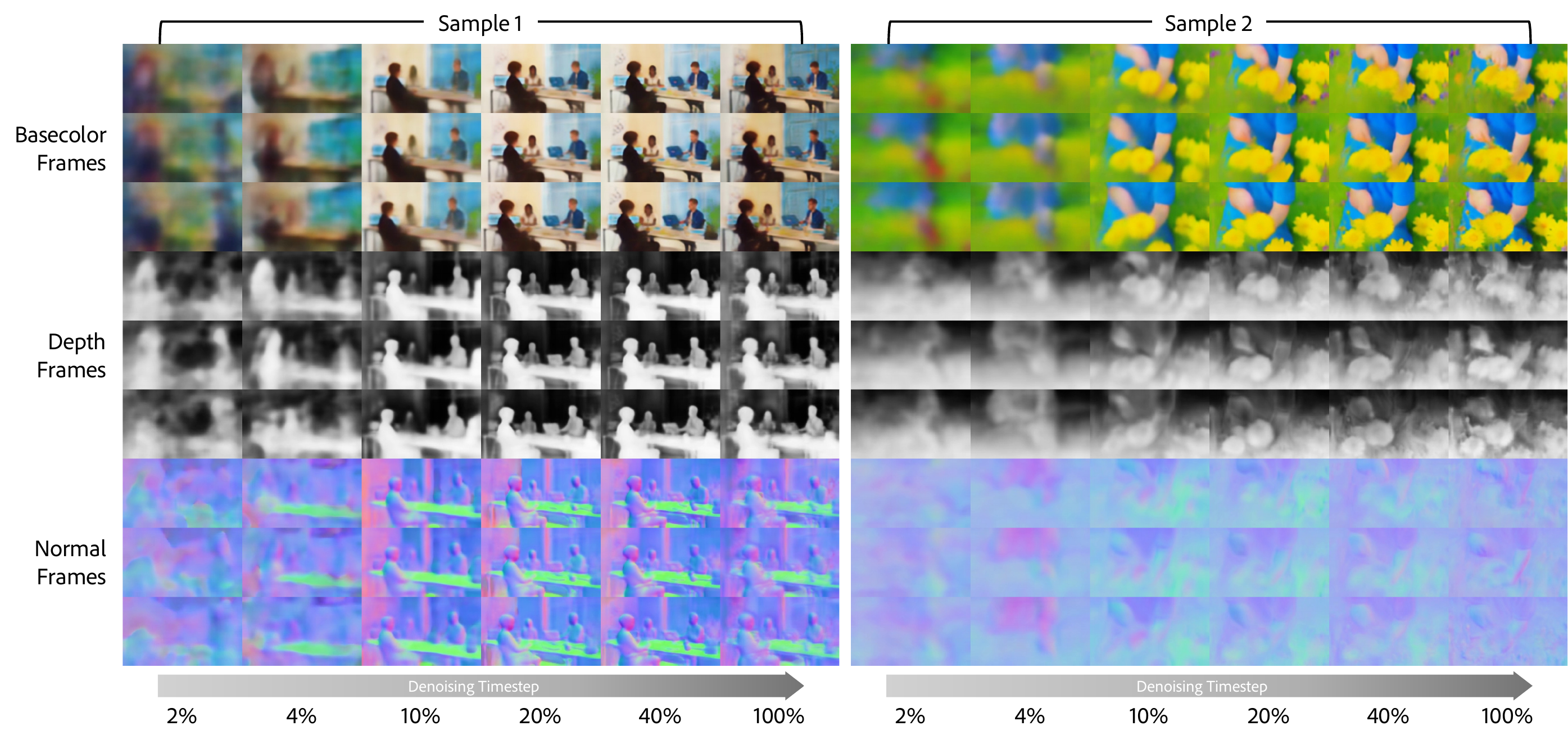}
\vspace{-10pt}
\caption{Timestep-wise evolution of base color, normal, and albedo. Coarse geometry appears early and refines through denoising.}
\vspace{-10pt}
\label{fig:timesteps-sup}
\end{figure*}

\begin{figure*}[!ht]
\centering
\includegraphics[width=1.0\linewidth]{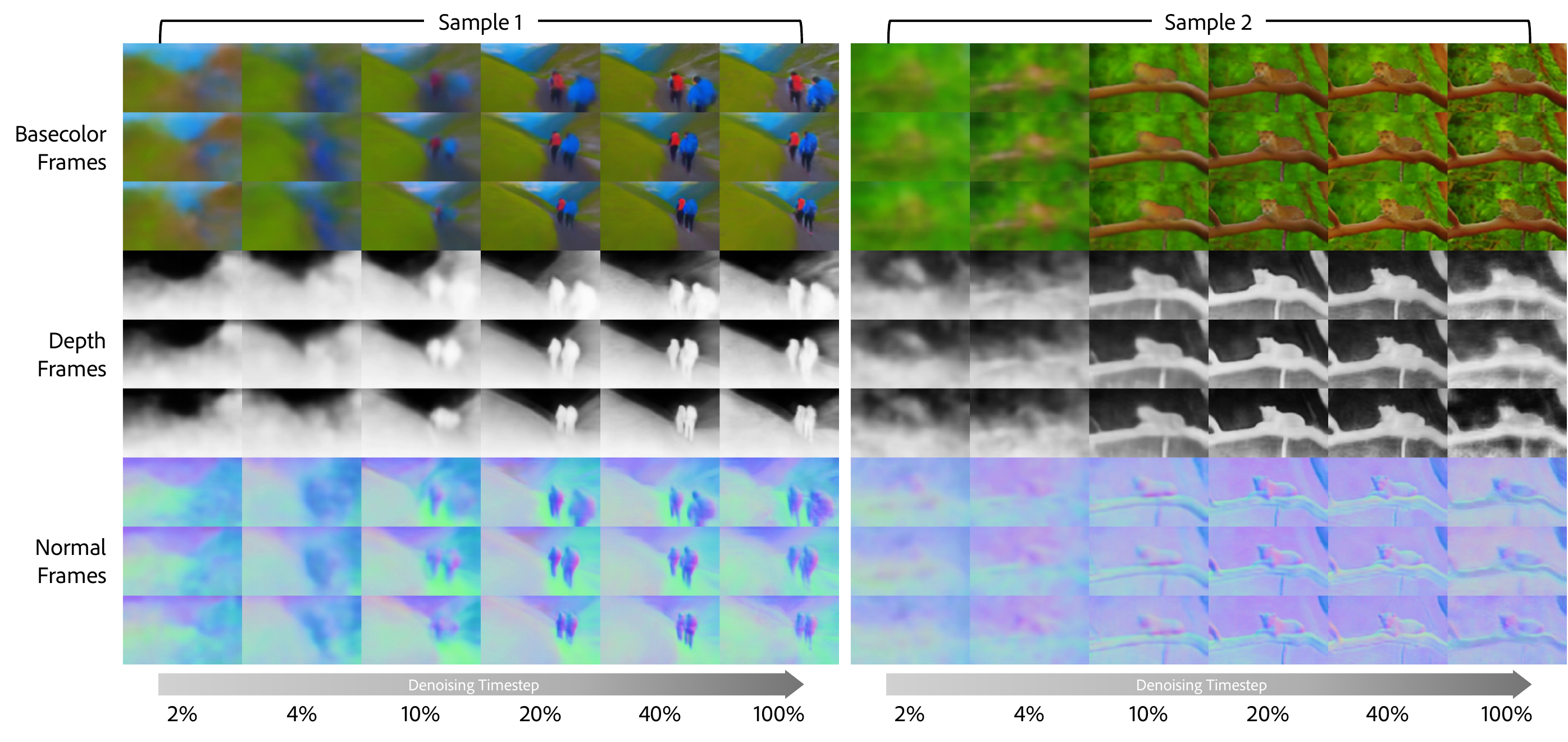}\\\vspace{20pt}
\includegraphics[width=1.0\linewidth]{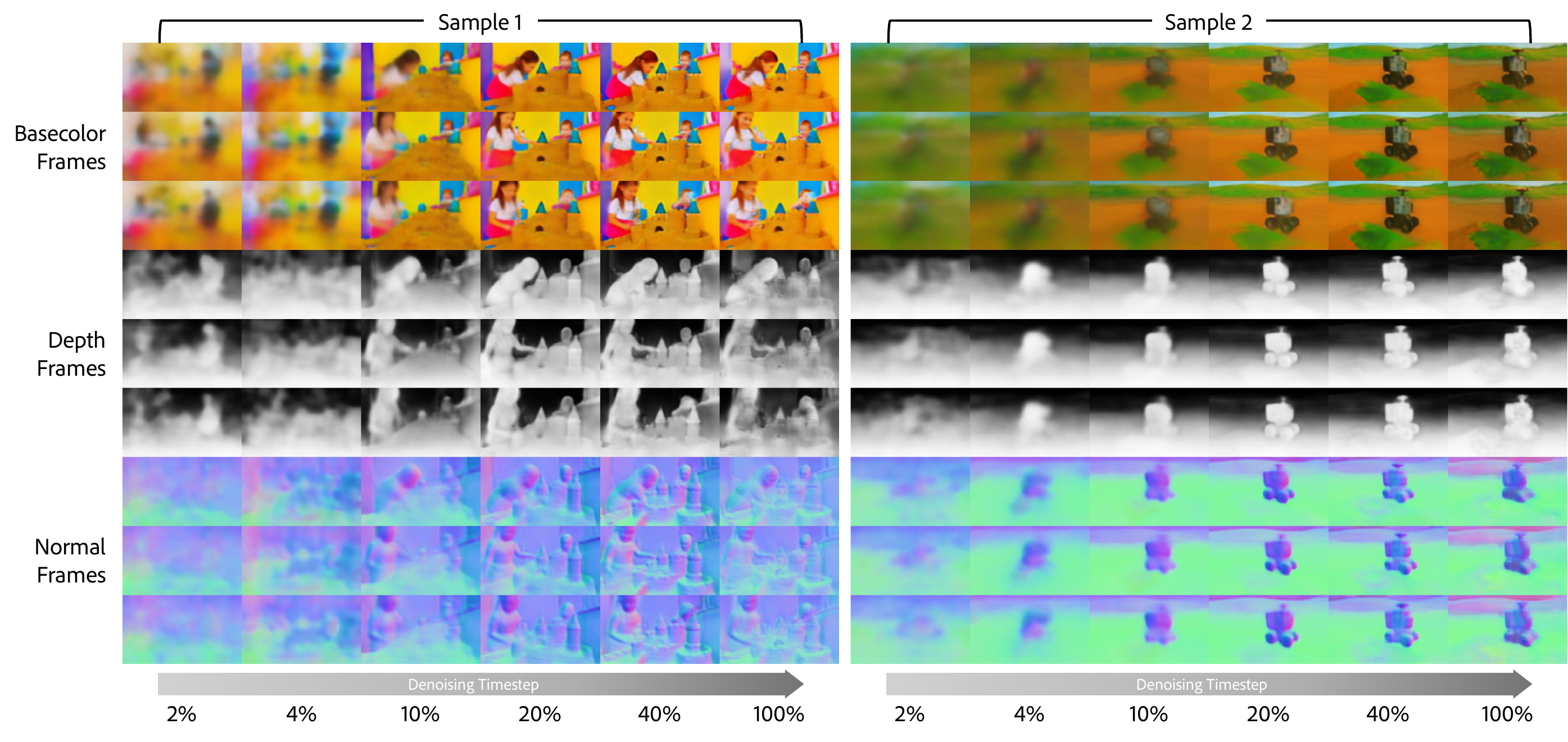}
\vspace{-10pt}
\caption{Timestep-wise evolution of base color, normal, and albedo. Coarse geometry appears early and refines through denoising.}
\vspace{-10pt}
\label{fig:timesteps-sup2}
\end{figure*}

\begin{figure*}[!ht]
\centering
\includegraphics[width=1.0\linewidth]{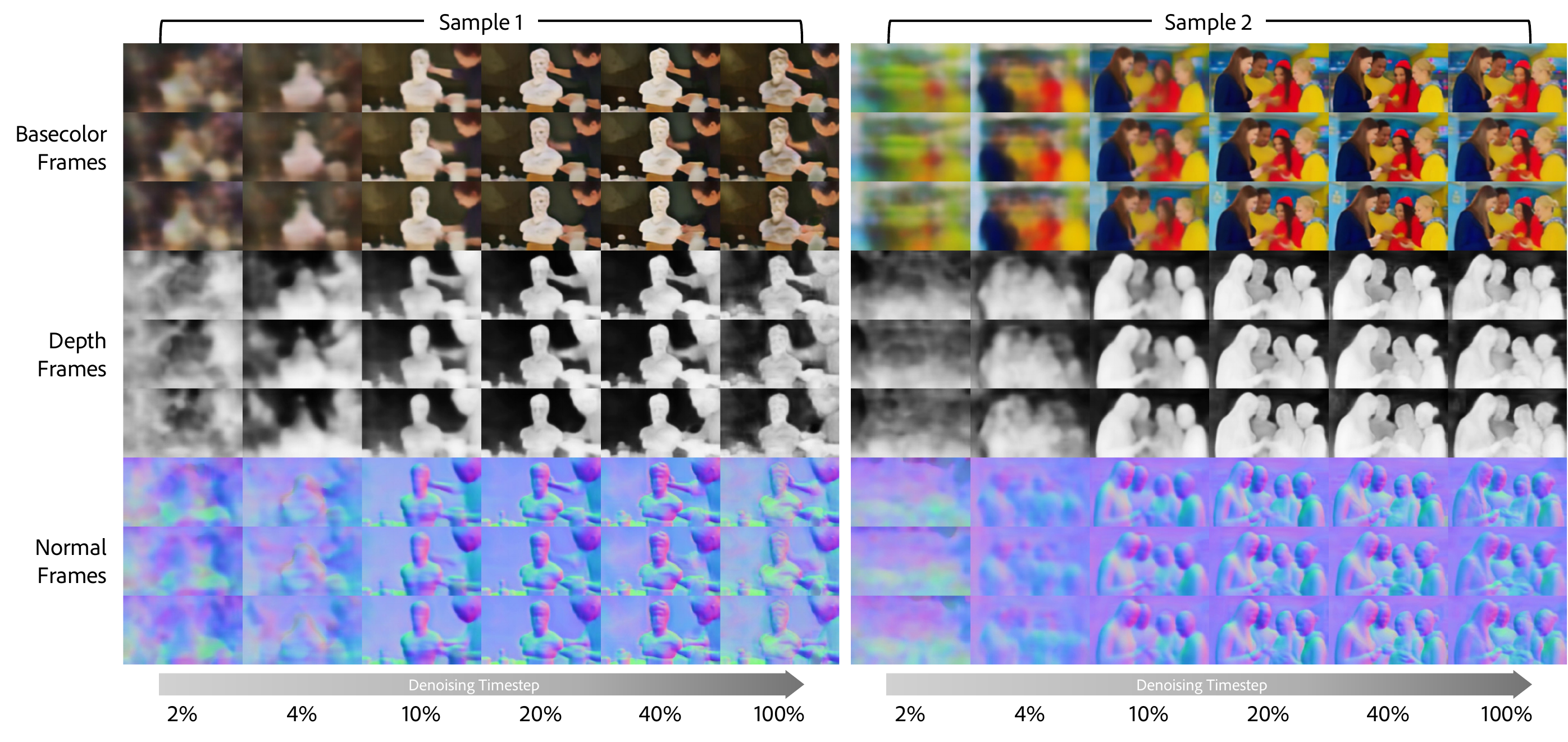}
\vspace{-10pt}
\caption{Timestep-wise evolution of base color, normal, and albedo. Coarse geometry appears early and refines through denoising.}
\vspace{-10pt}
\label{fig:timesteps-sup3}
\end{figure*}

Multi-modal previews provide several advantages over latent-space visualizations. First, intrinsic modalities, particularly depth and normals, reveal coarse scene geometry earlier in the denoising process than RGB or latents. Second, base color previews offer simplified appearance information without lighting, making scene layout clearer. Third, our method produces all previews simultaneously from the same features, allowing users to cross-reference modalities at any timestep. See Figures~\ref{fig:timesteps-sup}--\ref{fig:timesteps-sup3} for the timestep-wise evolution of the intrinsic modalities.

\begin{figure*}[t]
\centering
\includegraphics[width=0.25\linewidth]{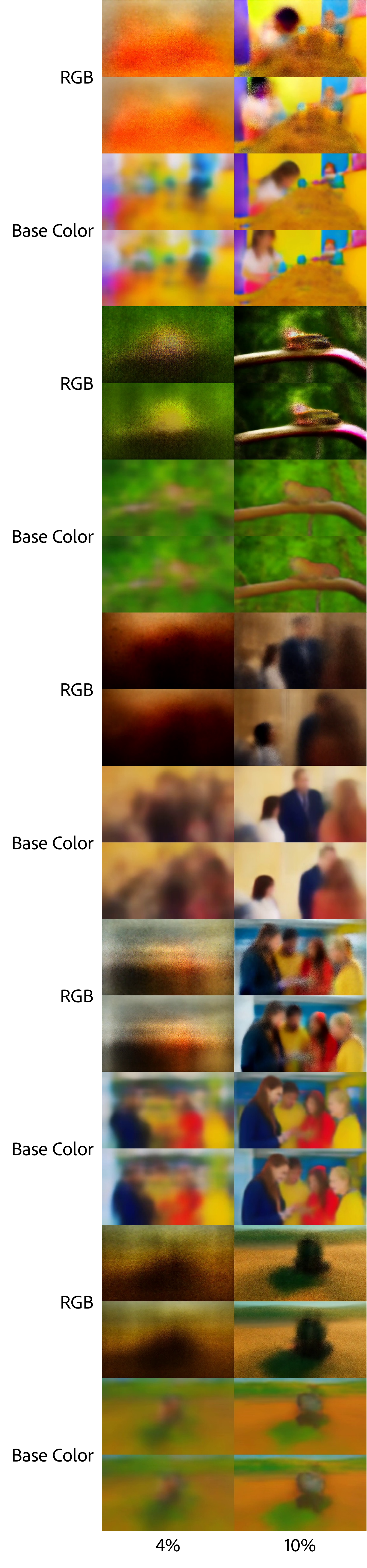}
\hspace{20pt}
\includegraphics[width=0.25\linewidth]{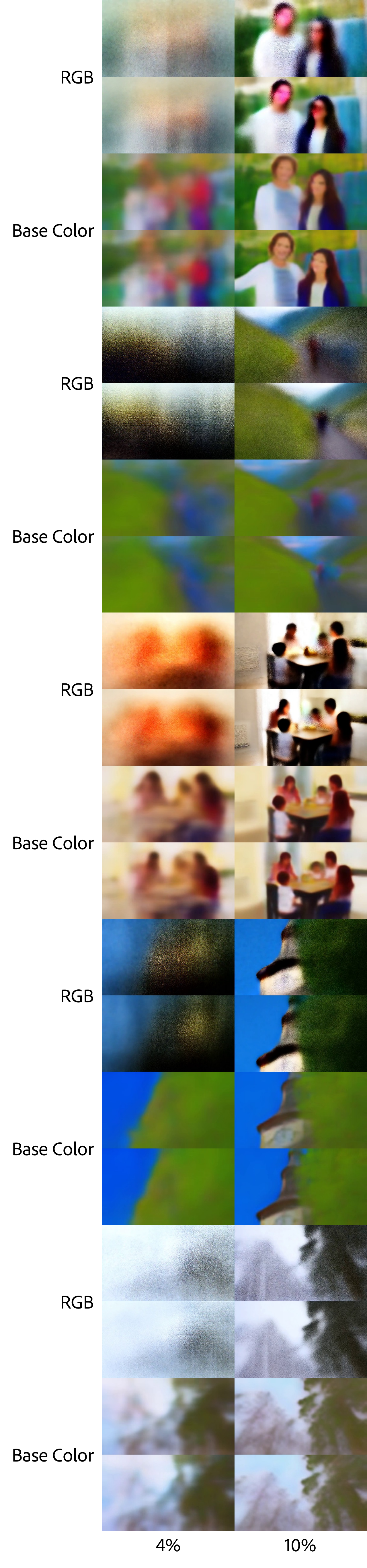}
\caption{Comparison between RGB (from \xzeropred) and base color (from our decoder) previews at 4\% and 10\% of the denoising steps. Our base color preview reveals structural components and color layout more clearly than the latent.}
\label{fig:intermediate-1}
\end{figure*}

Figure~\ref{fig:intermediate-1} compares latent renderings with our base color predictions. Intrinsic previews exhibit fewer lighting artifacts and present a cleaner structural representation during early timesteps.

\begin{table}[t]
\scriptsize
\centering
\begin{tabular}{llccc}
\toprule
\textbf{Data} & \textbf{Method} & \textbf{NFE} & \textbf{Mean \#Boxes} & \textbf{Std. \#Boxes} \\
\midrule
\multirow{6}{*}{\textbf{Only Motion}}
 & MB-Avg. & 1 & 4.0 & 0.2 \\
 & CD & 1 & 8.2 & 1.2 \\
 & DDPM & 20 & 7.5 & 1.1 \\
 & DDPM & 200 Prev. & 5.0 & 2.0 \\
 & DDPM & 1K & 4.0 & 0.0 \\
 & GT & - & 4.0 & 0.0 \\
\midrule
\multirow{5}{*}{\textbf{Motion+Position}}
 & MB-Avg. & 1 & 3.5 & 0.2 \\
 & CD & 1 & 0.4 & 0.5 \\
 & DDPM & 200 Prev. & 0.7 & 0.8 \\
 & DDPM & 1K & 3.8 & 0.4 \\
 & GT & - & 3.8 & 0.2 \\
\bottomrule
\end{tabular}
\caption{Toy experiment results across configurations. MB-Avg. denotes the mean across branches of the average and standard deviation of the number of boxes.}
\label{tab:toy}
\end{table}

\section{User Study Setup}
\label{sec:user_studies}

Here we provide additional details on the user study, specifically regarding the experimental setup and the questions participants were asked. The goal of the study was to evaluate the perceptual usefulness of intrinsic-based previews compared to the $\mathbf{x}_0$-pred baseline.

Each participant was shown two preview representations for a given reference video: (1) a standard $\mathbf{x}_0$-pred preview and (2) our intrinsic-based preview, where participants could additionally consult predicted modalities (e.g., base color, depth, normals) alongside the RGB preview. For each trial, participants answered three questions designed to measure complementary aspects of preview quality:
\begin{itemize}
    \item \textbf{Content Predictability:} ``Which representation allows you to predict the content of the reference video?" This measures how well a preview communicates the expected outcome of the diffusion process.
    \item \textbf{Visual Fidelity:} ``Which representation has fewer artifacts or errors (such as noise, flickering, etc.)?" This assesses perceived stability and cleanliness.
    \item \textbf{Scene Clarity:} ``Which video more clearly shows the scene composition (objects, motion, layout, etc.)?" This evaluates structural interpretability.
\end{itemize}

A total of 35 participants each evaluated 10 examples, yielding 350 responses for each question. As summarized in the main paper, participants consistently favored our intrinsic-based previews across all three criteria, indicating that intrinsic modalities provide more informative and reliable early-stage previews than the $\mathbf{x}_0$-pred baseline.

\begin{table*}[h]
\centering
\small
\begin{tabular}{clcl}
\toprule
\textbf{No.} & \textbf{Category} & \textbf{No.} & \textbf{Category} \\
\midrule
\multicolumn{4}{l}{\textit{Human-Focused Categories}} \\
1 & Human Portraits and Expressions & 2 & Human Daily Activities \\
3 & Sports and Physical Activities & 4 & Professions and Work Environments \\
5 & Human Emotions and Reactions & 6 & Celebrations and Social Events \\
\midrule
\multicolumn{4}{l}{\textit{Animal Categories}} \\
7 & Wild Animals in Natural Habitats & 8 & Domestic Animals and Pets \\
9 & Marine Life and Underwater Scenes & 10 & Birds and Flying Creatures \\
11 & Insects and Microscopic Life & & \\
\midrule
\multicolumn{4}{l}{\textit{Nature and Vegetation Categories}} \\
12 & Forests and Tree Scenes & 13 & Flowers and Garden Beauty \\
14 & Weather Phenomena & 15 & Natural Landscapes \\
16 & Seasonal Transformations & & \\
\midrule
\multicolumn{4}{l}{\textit{Indoor Environment Categories}} \\
17 & Home Interior Scenes & 18 & Restaurants and Dining \\
19 & Offices and Workspaces & 20 & Cultural and Educational Spaces \\
\midrule
\multicolumn{4}{l}{\textit{Outdoor Environment Categories}} \\
21 & Urban Cityscapes & 22 & Rural and Countryside \\
23 & Beach and Coastal Environments & 24 & Mountain and Adventure Scenes \\
\midrule
\multicolumn{4}{l}{\textit{Motion and Movement Categories}} \\
25 & Transportation and Vehicles & 26 & Dance and Choreography \\
27 & Flowing Elements (Water, Smoke, Particles) & 28 & Mechanical and Industrial Motion \\
\midrule
\multicolumn{4}{l}{\textit{Fantasy and Creative Categories}} \\
29 & Fantasy Creatures and Magic & 30 & Science Fiction and Futuristic \\
31 & Abstract and Surreal Concepts & 32 & Historical and Period Scenes \\
\midrule
\multicolumn{4}{l}{\textit{Complex Scene Categories}} \\
33 & Crowd Scenes and Gatherings & 34 & Action and Adventure \\
35 & Time-lapse and Slow Motion & 36 & Microscopic and Macro Worlds \\
\midrule
\multicolumn{4}{l}{\textit{Specialized Categories}} \\
37 & Artistic and Stylized Visuals & 38 & Cooking and Food Preparation \\
39 & Technology and Modern Gadgets & 40 & Transformations and Metamorphosis \\
\bottomrule
\end{tabular}
\caption{40 categories with 25 prompts each (1,000 total prompts)}
\label{tab:categories}
\end{table*}

\begin{table*}[!ht]
\centering
\small
\begin{tabular}{lccccccc}
\toprule
\textbf{Decoder Type} & \textbf{RGB} & \textbf{Base Color} & \textbf{Depth} & \textbf{Normal} & \textbf{Metallic} & \textbf{Roughness} \\
\midrule
\multicolumn{7}{l}{\textit{4\% Denoising Steps}} \\
\xzeropred & 12.07 & - & - & - & - & - \\
Linear & 14.74 & 13.88 & 14.52 & 18.92 & 10.98 & 15.56 \\
Ours & 14.75 & 14.16 & 14.86 & 19.02 & 15.53 & 16.15 \\
\midrule
\multicolumn{7}{l}{\textit{6\% Denoising Steps}} \\
\xzeropred & 14.15 & - & - & - & - & - \\
Linear & 16.17 & 14.74 & 15.10 & 19.17 & 11.02 & 15.69 \\
Ours & 16.05 & 15.22 & 15.74 & 19.39 & 16.43 & 16.60 \\
\midrule
\multicolumn{7}{l}{\textit{8\% Denoising Steps}} \\
\xzeropred & 15.65 & - & - & - & - & - \\
Linear & 17.19 & 15.20 & 15.35 & 19.39 & 11.19 & 15.67 \\
Ours & 17.16 & 15.89 & 16.52 & 19.66 & 16.05 & 16.74 \\
\midrule
\multicolumn{7}{l}{\textit{10\% Denoising Steps}} \\
\xzeropred & 16.98 & - & - & - & - & - \\
Linear & 17.96 & 15.51 & 15.52 & 19.54 & 11.22 & 15.75 \\
Ours & 18.03 & 16.38 & 16.95 & 20.04 & 16.42 & 17.03 \\
\midrule
\multicolumn{7}{l}{\textit{12\% Denoising Steps}} \\
\xzeropred & 18.01 & - & - & - & - & - \\
Linear & 18.50 & 15.68 & 15.62 & 19.63 & 11.23 & 15.75 \\
Ours & 18.70 & 16.79 & 17.28 & 20.11 & 15.95 & 17.35 \\
\midrule
\multicolumn{7}{l}{\textit{14\% Denoising Steps}} \\
\xzeropred & 18.83 & - & - & - & - & - \\
Linear & 18.88 & 15.80 & 15.68 & 19.70 & 11.25 & 15.75 \\
Ours & 19.22 & 17.00 & 17.59 & 20.31 & 17.10 & 17.27 \\
\midrule
\multicolumn{7}{l}{\textit{16\% Denoising Steps}} \\
\xzeropred & 19.59 & - & - & - & - & - \\
Linear & 19.18 & 15.88 & 15.69 & 19.76 & 11.25 & 15.75 \\
Ours & 19.58 & 17.08 & 17.70 & 20.44 & 16.83 & 17.49 \\
\midrule
\multicolumn{7}{l}{\textit{18\% Denoising Steps}} \\
\xzeropred & 20.16 & - & - & - & - & - \\
Linear & 19.41 & 15.94 & 15.69 & 19.79 & 11.25 & 15.73 \\
Ours & 19.92 & 17.29 & 17.71 & 20.56 & 17.14 & 17.47 \\
\midrule
\multicolumn{7}{l}{\textit{20\% Denoising Steps}} \\
\xzeropred & 20.69 & - & - & - & - & - \\
Linear & 19.60 & 15.97 & 15.69 & 19.82 & 11.25 & 15.71 \\
Ours & 20.19 & 17.34 & 17.86 & 20.52 & 16.83 & 17.73 \\
\bottomrule
\end{tabular}
\caption{PSNR results for different decoder types across denoising timesteps. Higher is better.}
\label{tab:psnr-decoder}
\end{table*}

\begin{table*}[!ht]
\centering
\small
\begin{tabular}{lccccccc}
\toprule
\textbf{Decoder Type} & \textbf{RGB} & \textbf{Base Color} & \textbf{Depth} & \textbf{Normal} & \textbf{Metallic} & \textbf{Roughness} \\
\midrule
\multicolumn{7}{l}{\textit{4\% Denoising Steps}} \\
\xzeropred & 0.069 & - & - & - & - & - \\
Linear & 0.037 & 0.047 & 0.044 & 0.017 & 0.143 & 0.042 \\
Ours & 0.037 & 0.046 & 0.046 & 0.016 & 0.155 & 0.042 \\
\midrule
\multicolumn{7}{l}{\textit{6\% Denoising Steps}} \\
\xzeropred & 0.043 & - & - & - & - & - \\
Linear & 0.027 & 0.038 & 0.038 & 0.016 & 0.132 & 0.040 \\
Ours & 0.029 & 0.037 & 0.037 & 0.015 & 0.127 & 0.039 \\
\midrule
\multicolumn{7}{l}{\textit{8\% Denoising Steps}} \\
\xzeropred & 0.031 & - & - & - & - & - \\
Linear & 0.022 & 0.035 & 0.036 & 0.015 & 0.130 & 0.040 \\
Ours & 0.022 & 0.031 & 0.032 & 0.014 & 0.127 & 0.038 \\
\midrule
\multicolumn{7}{l}{\textit{10\% Denoising Steps}} \\
\xzeropred & 0.024 & - & - & - & - & - \\
Linear & 0.018 & 0.032 & 0.034 & 0.014 & 0.128 & 0.039 \\
Ours & 0.019 & 0.028 & 0.029 & 0.013 & 0.131 & 0.034 \\
\midrule
\multicolumn{7}{l}{\textit{12\% Denoising Steps}} \\
\xzeropred & 0.019 & - & - & - & - & - \\
Linear & 0.016 & 0.031 & 0.034 & 0.014 & 0.128 & 0.038 \\
Ours & 0.016 & 0.026 & 0.027 & 0.012 & 0.133 & 0.033 \\
\midrule
\multicolumn{7}{l}{\textit{14\% Denoising Steps}} \\
\xzeropred & 0.016 & - & - & - & - & - \\
Linear & 0.015 & 0.031 & 0.033 & 0.014 & 0.128 & 0.038 \\
Ours & 0.014 & 0.025 & 0.025 & 0.012 & 0.126 & 0.034 \\
\midrule
\multicolumn{7}{l}{\textit{16\% Denoising Steps}} \\
\xzeropred & 0.014 & - & - & - & - & - \\
Linear & 0.014 & 0.030 & 0.033 & 0.013 & 0.127 & 0.038 \\
Ours & 0.013 & 0.024 & 0.025 & 0.011 & 0.126 & 0.032 \\
\midrule
\multicolumn{7}{l}{\textit{18\% Denoising Steps}} \\
\xzeropred & 0.012 & - & - & - & - & - \\
Linear & 0.013 & 0.030 & 0.033 & 0.013 & 0.127 & 0.038 \\
Ours & 0.012 & 0.024 & 0.025 & 0.011 & 0.121 & 0.032 \\
\midrule
\multicolumn{7}{l}{\textit{20\% Denoising Steps}} \\
\xzeropred & 0.011 & - & - & - & - & - \\
Linear & 0.013 & 0.030 & 0.033 & 0.013 & 0.127 & 0.038 \\
Ours & 0.012 & 0.023 & 0.024 & 0.011 & 0.123 & 0.033 \\
\bottomrule
\end{tabular}
\caption{MSE results for different decoder types across denoising timesteps. Lower is better.}
\label{tab:mse-decoder}
\end{table*}

\begin{table*}[!ht]
\centering
\small
\begin{tabular}{lccccccc}
\toprule
\textbf{Decoder Type} & \textbf{RGB} & \textbf{Base Color} & \textbf{Depth} & \textbf{Normal} & \textbf{Metallic} & \textbf{Roughness} \\
\midrule
\multicolumn{7}{l}{\textit{4\% Denoising Steps}} \\
\xzeropred & 0.186 & - & - & - & - & - \\
Linear & 0.141 & 0.166 & 0.161 & 0.091 & 0.309 & 0.168 \\
Ours & 0.135 & 0.156 & 0.148 & 0.090 & 0.271 & 0.160 \\
\midrule
\multicolumn{7}{l}{\textit{6\% Denoising Steps}} \\
\xzeropred & 0.140 & - & - & - & - & - \\
Linear & 0.115 & 0.148 & 0.149 & 0.088 & 0.297 & 0.163 \\
Ours & 0.112 & 0.136 & 0.133 & 0.085 & 0.241 & 0.152 \\
\midrule
\multicolumn{7}{l}{\textit{8\% Denoising Steps}} \\
\xzeropred & 0.114 & - & - & - & - & - \\
Linear & 0.099 & 0.140 & 0.143 & 0.085 & 0.293 & 0.163 \\
Ours & 0.095 & 0.124 & 0.121 & 0.081 & 0.242 & 0.149 \\
\midrule
\multicolumn{7}{l}{\textit{10\% Denoising Steps}} \\
\xzeropred & 0.093 & - & - & - & - & - \\
Linear & 0.090 & 0.134 & 0.140 & 0.084 & 0.291 & 0.161 \\
Ours & 0.084 & 0.117 & 0.115 & 0.078 & 0.241 & 0.142 \\
\midrule
\multicolumn{7}{l}{\textit{12\% Denoising Steps}} \\
\xzeropred & 0.080 & - & - & - & - & - \\
Linear & 0.084 & 0.132 & 0.139 & 0.083 & 0.291 & 0.161 \\
Ours & 0.077 & 0.111 & 0.111 & 0.077 & 0.243 & 0.139 \\
\midrule
\multicolumn{7}{l}{\textit{14\% Denoising Steps}} \\
\xzeropred & 0.072 & - & - & - & - & - \\
Linear & 0.080 & 0.130 & 0.138 & 0.082 & 0.290 & 0.160 \\
Ours & 0.072 & 0.109 & 0.107 & 0.075 & 0.233 & 0.140 \\
\midrule
\multicolumn{7}{l}{\textit{16\% Denoising Steps}} \\
\xzeropred & 0.065 & - & - & - & - & - \\
Linear & 0.077 & 0.129 & 0.138 & 0.082 & 0.290 & 0.160 \\
Ours & 0.069 & 0.108 & 0.107 & 0.074 & 0.234 & 0.135 \\
\midrule
\multicolumn{7}{l}{\textit{18\% Denoising Steps}} \\
\xzeropred & 0.060 & - & - & - & - & - \\
Linear & 0.075 & 0.129 & 0.138 & 0.081 & 0.290 & 0.160 \\
Ours & 0.065 & 0.106 & 0.106 & 0.073 & 0.230 & 0.136 \\
\midrule
\multicolumn{7}{l}{\textit{20\% Denoising Steps}} \\
\xzeropred & 0.056 & - & - & - & - & - \\
Linear & 0.073 & 0.128 & 0.138 & 0.081 & 0.290 & 0.160 \\
Ours & 0.063 & 0.105 & 0.104 & 0.073 & 0.230 & 0.135 \\
\bottomrule
\end{tabular}
\caption{L1 error results for different decoder types across denoising timesteps. Lower is better.}
\label{tab:l1-decoder}
\end{table*}

\begin{table*}[!ht]
\centering
\small
\begin{tabular}{lccccccc}
\toprule
\textbf{Decoder Type} & \textbf{RGB} & \textbf{Base Color} & \textbf{Depth} & \textbf{Normal} & \textbf{Metallic} & \textbf{Roughness} \\
\midrule
\multicolumn{7}{l}{\textit{4\% Denoising Steps}} \\
\xzeropred & 0.751 & - & - & - & - & - \\
Linear & 0.706 & 0.681 & 0.665 & 0.505 & 0.791 & 0.683 \\
Ours & 0.689 & 0.601 & 0.501 & 0.456 & 0.607 & 0.545 \\
\midrule
\multicolumn{7}{l}{\textit{6\% Denoising Steps}} \\
\xzeropred & 0.634 & - & - & - & - & - \\
Linear & 0.628 & 0.623 & 0.651 & 0.495 & 0.789 & 0.689 \\
Ours & 0.601 & 0.528 & 0.467 & 0.423 & 0.561 & 0.515 \\
\midrule
\multicolumn{7}{l}{\textit{8\% Denoising Steps}} \\
\xzeropred & 0.522 & - & - & - & - & - \\
Linear & 0.571 & 0.582 & 0.641 & 0.483 & 0.787 & 0.692 \\
Ours & 0.523 & 0.463 & 0.435 & 0.393 & 0.556 & 0.497 \\
\midrule
\multicolumn{7}{l}{\textit{10\% Denoising Steps}} \\
\xzeropred & 0.433 & - & - & - & - & - \\
Linear & 0.535 & 0.554 & 0.632 & 0.469 & 0.785 & 0.692 \\
Ours & 0.466 & 0.422 & 0.411 & 0.362 & 0.538 & 0.477 \\
\midrule
\multicolumn{7}{l}{\textit{12\% Denoising Steps}} \\
\xzeropred & 0.369 & - & - & - & - & - \\
Linear & 0.511 & 0.537 & 0.627 & 0.460 & 0.785 & 0.692 \\
Ours & 0.426 & 0.386 & 0.392 & 0.346 & 0.536 & 0.465 \\
\midrule
\multicolumn{7}{l}{\textit{14\% Denoising Steps}} \\
\xzeropred & 0.319 & - & - & - & - & - \\
Linear & 0.494 & 0.525 & 0.624 & 0.453 & 0.784 & 0.693 \\
Ours & 0.397 & 0.366 & 0.381 & 0.333 & 0.519 & 0.462 \\
\midrule
\multicolumn{7}{l}{\textit{16\% Denoising Steps}} \\
\xzeropred & 0.279 & - & - & - & - & - \\
Linear & 0.482 & 0.517 & 0.621 & 0.447 & 0.783 & 0.692 \\
Ours & 0.373 & 0.350 & 0.376 & 0.325 & 0.517 & 0.453 \\
\midrule
\multicolumn{7}{l}{\textit{18\% Denoising Steps}} \\
\xzeropred & 0.249 & - & - & - & - & - \\
Linear & 0.472 & 0.510 & 0.620 & 0.444 & 0.783 & 0.691 \\
Ours & 0.354 & 0.338 & 0.368 & 0.316 & 0.508 & 0.451 \\
\midrule
\multicolumn{7}{l}{\textit{20\% Denoising Steps}} \\
\xzeropred & 0.224 & - & - & - & - & - \\
Linear & 0.464 & 0.505 & 0.619 & 0.441 & 0.782 & 0.690 \\
Ours & 0.343 & 0.331 & 0.362 & 0.314 & 0.511 & 0.445 \\
\bottomrule
\end{tabular}
\caption{LPIPS results for different decoder types across denoising timesteps. Lower is better.}
\label{tab:lpips-decoder}
\end{table*}

\clearpage

\clearpage

\clearpage

\end{document}